\documentclass{article}

\PassOptionsToPackage{numbers,compress}{natbib}
\usepackage[preprint]{neurips_2026}

\usepackage[utf8]{inputenc} 
\usepackage[T1]{fontenc}    
\usepackage{hyperref}       
\usepackage{url}            
\usepackage{booktabs}       
\usepackage{amsfonts}       
\usepackage{nicefrac}       
\usepackage{microtype}      
\usepackage{xcolor}         

\usepackage{amsmath}
\usepackage{amssymb}
\usepackage{graphicx}
\usepackage{wrapfig}
\usepackage{caption}
\usepackage{subcaption}
\usepackage{booktabs}
\usepackage{multirow}
\usepackage{array}
\usepackage{pifont}
\usepackage{enumitem}
\usepackage{algorithm}
\usepackage{algorithmic}
\usepackage{wrapfig}
\usepackage{float}

\newcommand{\xmark}{\ding{55}}

\usepackage[table]{xcolor}

\definecolor{mydarkred}{RGB}{186,49,49}
\definecolor{mydarkblue}{RGB}{7,108,186}
\definecolor{lightblue}{RGB}{203,240,246}

\title{DIMoE-Adapters: Dynamic Expert Evolution for Continual Learning in Vision--Language Models}

\author{%
Mengxin Qin \quad Xiang Zhang \quad Xi Wang \quad Kun Wei\thanks{Corresponding author.} \quad Xu Yang \quad Cheng Deng\\
School of Electronic Engineering, Xidian University Xi'an 710071, China\\
}

\begin{document}

\maketitle

\begin{abstract}
Continual learning (CL) enables Vision--Language Models (VLMs) to accumulate knowledge and adapt to evolving tasks without the need for retraining from scratch.
However, in the context of Multi-domain Task-Incremental Learning (MTIL), significant domain shifts exacerbate the stability–plasticity dilemma.
Most existing methods rely on fixed architectures with statically allocated parameters, constraining adaptation to new domains and aggravating catastrophic forgetting.
To address these challenges, we propose DIMoE-Adapters (\textbf{D}ynamic \textbf{I}ncremental \textbf{M}ixture-\textbf{o}f-\textbf{E}xperts \textbf{Adapters}), a framework that introduces a Dynamic Expert Evolution paradigm to balance stability and plasticity.
This paradigm is realized through two collaborative components: Self-Calibrated Expert Evolution (SCEE) and Prototype-Guided Expert Selection (PGES).
Specifically, SCEE constructs and evolves a sparse expert pool through expert optimization dynamics, thereby enhancing plasticity while minimizing redundant capacity.
PGES governs expert utilization based on the pool shaped by SCEE, improving model stability across previously encountered and unseen tasks.
Extensive experiments show that our approach outperforms previous state-of-the-art methods across various settings.
\end{abstract}

\section{Introduction}
Continual learning (CL)~\cite{parisi2019continual,de2021continual,buzzega2020dark} aims to enable models to acquire new knowledge incrementally while reducing catastrophic forgetting of previously learned information, a challenge known as the stability--plasticity dilemma~\cite{zenke2017continual,shin2017continual,tiwari2022gcr}.
With the emergence of Vision--Language Models (VLMs) such as CLIP~\cite{radford2021learning}, the focus of CL has shifted from training models from scratch to adapting pretrained representations~\cite{zhou2022learning}.
Although pretrained VLMs exhibit strong generalization, preserving it during continual adaptation further exacerbates the stability--plasticity trade-off.

This challenge becomes particularly severe in Multi-domain Task-Incremental Learning (MTIL)~\cite{zheng2023preventing}, where tasks arrive sequentially from different visual domains, causing distribution shifts.
Adapting to such shifts requires strong plasticity to capture domain-specific patterns, yet aggressive updates often undermine stability in two critical aspects:
(\emph{i}) performance on previously learned tasks deteriorates due to catastrophic forgetting, and
(\emph{ii}) pretrained vision--language alignment is disrupted, leading to degraded zero-shot generalization.
Therefore, the central challenge in MTIL lies in balancing plasticity for new tasks and stability for old knowledge while preserving zero-shot generalization.

\begin{wrapfigure}{r}{0.48\textwidth}
    \vspace{-10pt}
    \centering
    \includegraphics[width=0.95\linewidth]{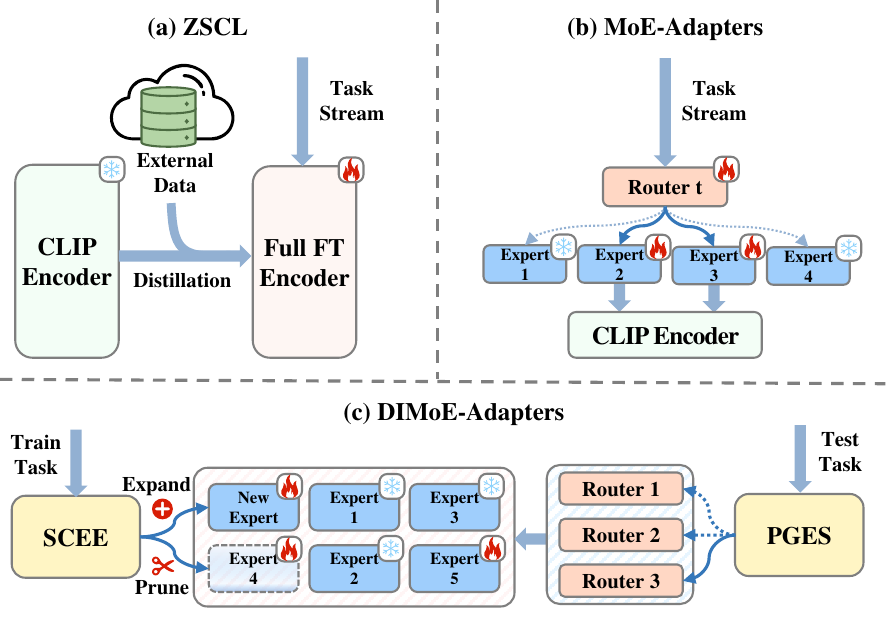}
    \caption{
    Comparison of three paradigms for continual learning of vision--language models: 
    (a) ZSCL~\cite{zheng2023preventing}, 
    (b) MoE-Adapters~\cite{yu2024boosting}, 
    and (c) DIMoE-Adapters.
    }
    \vspace{-10pt}
    \label{fig:small-framework}
\end{wrapfigure}

Most existing methods struggle to achieve this balance.
Early approaches like ZSCL~\cite{zheng2023preventing}, as shown in \autoref{fig:small-framework}(a), reduce forgetting via knowledge distillation with external data and parameter regularization, but incur high data and computational costs.
Recent parameter-efficient fine-tuning approaches like MoE-Adapters~\cite{yu2024boosting}, as illustrated in \autoref{fig:small-framework}(b), reduce interference by assigning task-specific knowledge to different experts.
Despite their effectiveness, these methods rely on fixed architectures with statically allocated parameter capacity, resulting in an inflexible stability--plasticity trade-off that limits long-term continual learning.

Recent dynamic expansion methods attempt to address this limitation by enabling adaptive model growth. For instance, SEMA~\cite{wang2025self} expands task-specific modules upon detecting distribution shifts, improving plasticity. However, it relies on additional detection modules to trigger expansion, increasing complexity and making expansion decisions sensitive to detection errors.

To overcome these limitations, continual learning in VLMs requires a dynamic expert management mechanism that jointly governs both expert evolution and utilization across tasks.
Therefore, we propose DIMoE-Adapters (\autoref{fig:small-framework}(c)), a dynamic mixture-of-experts framework that introduces a Dynamic Expert Evolution paradigm to balance stability and plasticity through Self-Calibrated Expert Evolution (SCEE) and Prototype-Guided Expert Selection (PGES).

SCEE is an optimization-driven strategy designed to enhance model plasticity. It monitors expert optimization dynamics and activation behaviors during training, enabling adaptive expansion, pruning, and freezing of experts.
Building upon the expert pool shaped by SCEE, PGES governs expert utilization at inference via prototype-guided task identification, ensuring stable performance on seen tasks while preserving CLIP’s zero-shot generalization on unseen ones.

Our contributions are summarized as follows:

\begin{itemize}
  \item We propose a Dynamic Expert Evolution paradigm for continual learning of VLMs, addressing the stability--plasticity dilemma in MTIL.
  \item We design DIMoE-Adapters, a framework that integrates SCEE for expert evolution during training and PGES for stable expert utilization at inference.
  \item Extensive experiments on MTIL and standard CL benchmarks demonstrate that DIMoE-Adapters consistently outperforms existing state-of-the-art methods in both forgetting mitigation and zero-shot preservation.
\end{itemize}

\section{Related Work}
\paragraph{Continual Learning.}
Existing continual learning methods can be broadly categorized into regularization-based, architecture-based, and replay-based approaches. 
Regularization-based methods~\cite{wei2021incremental,zenke2017continual} constrain parameter updates by importance estimation, but may overly restrict model plasticity.
Architecture-based methods~\cite{rusu2016progressive,mallya2018packnet} allocate task-specific parameters to reduce forgetting, but may hinder knowledge transfer across tasks.
Replay-based methods~\cite{wei2025compress,lopez2017gradient} mitigate forgetting by rehearsing stored or generated samples. 
However, most existing approaches focus on a single incremental setting, such as class-incremental~\cite{rebuffi2017icarl} or domain-incremental~\cite{volpi2018generalizing} learning, which limits their applicability to complex scenarios. 
In contrast, this work investigates catastrophic forgetting under the more challenging Multi-domain Task-Incremental Learning setting~\cite{zheng2023preventing}, which better reflects practical real-world applications.

\paragraph{Parameter-Efficient Fine-Tuning.}
Parameter-efficient fine-tuning (PEFT) adapts large pretrained models by introducing a small set of trainable parameters while keeping the majority of pretrained weights fixed, thereby avoiding the high computational cost and optimization instability associated with full fine-tuning.
Representative techniques include prompt learning~\cite{jia2022visual,liu2023pre}, adapters~\cite{houlsby2019parameter,wang2021k}, and low-rank adaptation (LoRA)~\cite{hu2022lora}, which have demonstrated strong performance in adapting foundation models to downstream tasks with minimal parameter overhead.
Recently, several CLIP-based continual learning methods have incorporated PEFT to alleviate catastrophic forgetting by selectively updating task-specific parameters or freezing subsets of the model~\cite{zhou2022learning,jha2024clap4clip}.
These approaches are primarily designed for class-incremental learning, emphasizing knowledge retention on previously learned tasks while placing less emphasis on preserving CLIP’s zero-shot generalization~\cite{buzzega2020dark,wang2022dualprompt}.
In this work, we propose a PEFT-based continual learning framework that jointly addresses catastrophic forgetting and zero-shot generalization.

\paragraph{Mixture-of-Experts.}
Mixture-of-Experts (MoE)~\cite{jacobs1991adaptive,shazeer2017outrageously} models comprise multiple expert modules coordinated by a routing mechanism that selectively assigns inputs to experts and aggregates their outputs.
Their modular structure has been used in continual learning to reduce cross-task interference~\cite{aljundi2017expert,chen2023lifelong}.
Prior work~\cite{yu2024boosting} explores task-specific experts with gating mechanisms to improve knowledge isolation and reuse. 
These approaches suggest that MoE can be an effective paradigm for mitigating catastrophic forgetting in continual learning.
Following this line of research, we propose a dynamic incremental MoE adapter framework, where adapters serve as experts and a Dynamic Expert Evolution paradigm adjusts expert behaviors to balance stability and plasticity.

\section{Methodology}
\subsection{Problem Definition}
We consider the Multi-domain Task-Incremental Learning (MTIL) setting~\cite{zheng2023preventing}, where a model learns from a sequence of $T$ tasks $\{\tau^{(1)}, \dots, \tau^{(T)}\}$. At each step $t$, the current task $\tau^{(t)}$ provides a dataset $\mathcal{S}^{(t)} = {(x_k^{(t)}, y_k^{(t)})}_{k=1}^{N_t}$ with inputs $x_k^{(t)} \in \mathcal{X}^{(t)}$ and labels $y_k^{(t)} \in \mathcal{Y}^{(t)}$. MTIL is characterized by disjoint label spaces ($\mathcal{Y}^{(i)} \cap \mathcal{Y}^{(j)} = \emptyset$ for $i \neq j$) and significant domain shifts ($\mathcal{D}^{(i)} \neq \mathcal{D}^{(j)}$). The goal is to maintain strong performance on all seen tasks $\{\tau^{(1)}, \dots, \tau^{(t)}\}$ while preserving zero-shot generalization to unseen tasks, without access to task identities during inference.

\subsection{Dynamic Incremental Mixture-of-Experts Adapters}
\begin{figure*}[t]
  \begin{center}
    \centerline{\includegraphics[width=0.95\textwidth]{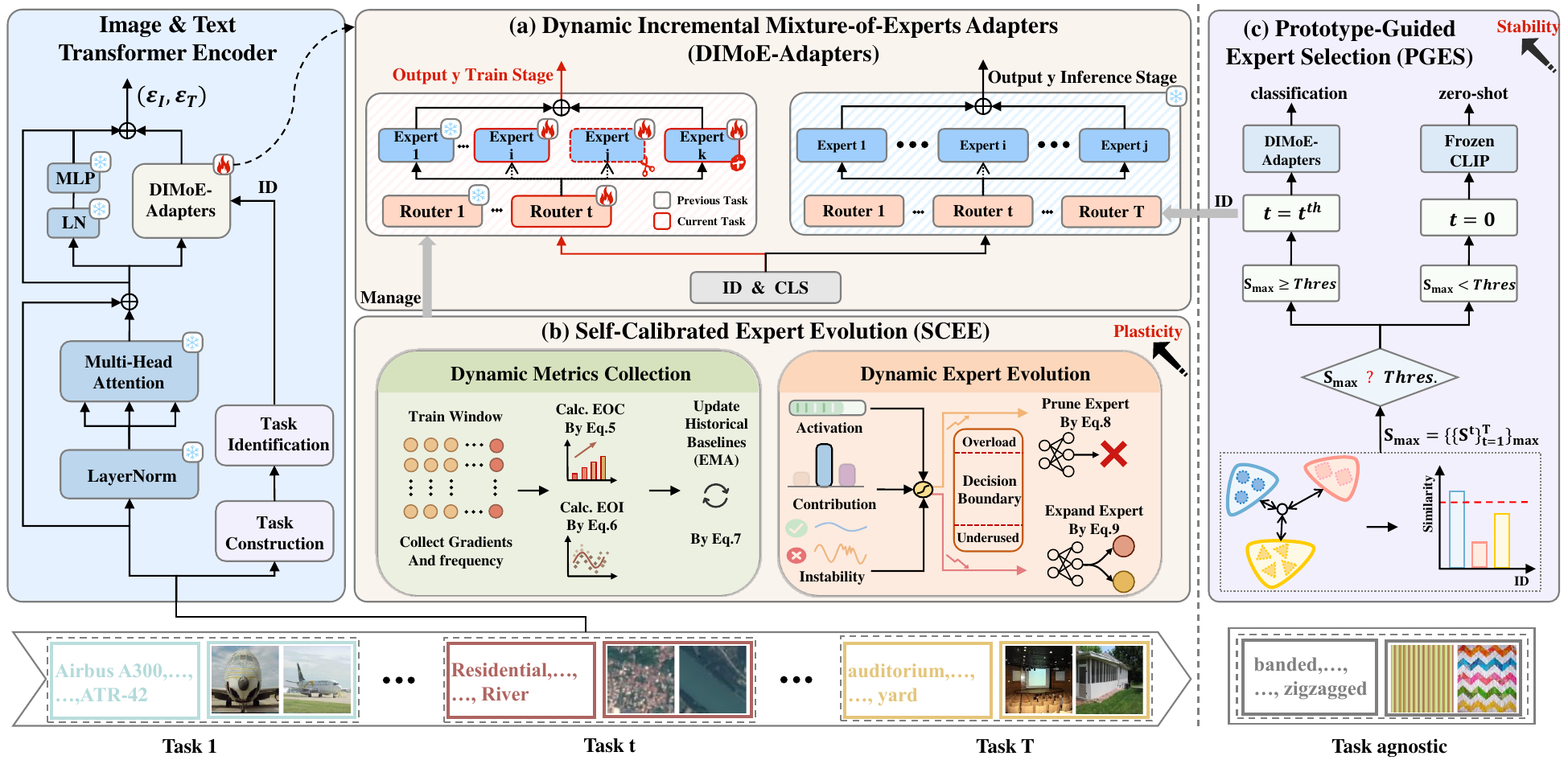}}
    \caption{
Overall framework of the proposed method.
(a) \textbf{DIMoE-Adapters.}
During training, a task-specific router and a set of experts are introduced to learn from $D_t$, while previously learned routers and experts are frozen. At inference, task-specific routers activate a sparse subset of experts for efficient prediction. (b) \textbf{SCEE.}
At the training stage, SCEE integrates Dynamic Metrics Collection and Dynamic Expert Evolution to manage the expert pool.
(c) \textbf{PGES.}
At the inference stage, PGES performs task identification via task-specific prototypes and routes samples to either DIMoE-Adapters for classification or the frozen CLIP model for zero-shot inference.
}
    \label{overall-framework}
  \end{center}
\end{figure*}

As shown in \autoref{overall-framework}, we propose DIMoE-Adapters, a dynamic mixture-of-experts framework for continual learning in VLMs.
Our framework adapts a frozen CLIP backbone~\cite{gao2024clip} by residually inserting Mixture-of-Experts (MoE) adapters into the feedforward layers.

At task $t$, we maintain a dynamically evolving expert pool
$\mathcal{E}^{(t)} = \{ E_i \}_{i=1}^{N_t}$, which contains frozen experts from previous tasks and newly introduced experts.
All experts are accessible through a task-specific router $\mathcal{R}^{(t)}$. Given an input $\mathbf{x}_\text{MoE} \in \mathbb{R}^D$, the MoE layer computes:
\begin{equation}
\mathbf{y} = \mathbf{x}_\text{MoE} + \sum_{i=1}^{N_t} w_i E_i(\mathbf{x}_\text{MoE}),
\label{eq:moe_output}
\end{equation}
where $w_i$ is the sparse routing weight assigned to expert $E_i$.

\paragraph{Top-$p$ Probability Accumulation Gating.}
We use the \texttt{[CLS]} token $\mathbf{c}$ from $\mathbf{x}_{\text{MoE}}$, obtained from its visual or textual token sequence, as the routing query and compute expert selection probabilities:
\begin{equation}
    \mathbf{p} = \text{Softmax}(\mathcal{R}^{(t)}(\mathbf{c})).
\end{equation}
We adopt an adaptive Top-$p$ gating strategy~\cite{huang2024harder} for expert selection.
The probabilities are sorted in descending order, and the smallest set $\mathcal{S}$ is selected such that their cumulative probability exceeds $p_0$ (default $0.6$).
The routing weights are re-normalized as:
\begin{equation}
    w_i = 
    \begin{cases} 
    \dfrac{p_i}{\sum_{j \in \mathcal{S}} p_j}, & i \in \mathcal{S}, \\
    0, & \text{otherwise}.
    \end{cases}
    \label{eq:renormalization}
\end{equation}

\paragraph{Task-Aware Load Balancing Loss.}
To further regulate expert utilization during continual training, we introduce a task-aware load balancing loss:
\begin{equation}
    \mathcal{L}_{\mathrm{balance}} = N_t \sum_{i=1}^{N_t} \beta_i \cdot (f_i \, Q_i),
\end{equation}
where $N_t$ denotes the number of experts at step $t$, $f_i$ denotes the fraction of tokens routed to expert $E_i$, and $Q_i$ is the average routing probability assigned to $E_i$, following prior MoE work~\cite{lepikhin2020gshard,fedus2022switch}.

Unlike standard load balancing objectives, we adapt the weighting coefficient $\beta_i$ according to expert type.
For experts inherited from previous tasks, we use $\beta_i = 1$ to maintain stable utilization.
For newly introduced experts in the current task, we set $\beta_i \in (0,1)$ (default $0.6$) to encourage exploration and promote their utilization during early adaptation.

\paragraph{Continual Training Protocol.}
We adopt a Freeze and Evolve training protocol.

\textbf{Phase 1: Freeze.}
Routers and experts associated with previously learned tasks are frozen to preserve knowledge, while the frozen experts remain in the expert pool for selection by the new router.

\textbf{Phase 2: Evolve.}
When a new task arrives, a task-specific router and a set of candidate experts are introduced to model the novel distribution.
Within a limited evolutionary window, these experts are governed by the proposed SCEE (\autoref{sec:SCEE}), which expands and prunes experts.

\subsection{Self-Calibrated Expert Evolution}
\label{sec:SCEE}
We introduce Self-Calibrated Expert Evolution (SCEE), a mechanism for controlling expert expansion and pruning during continual training.
For each task, expert behaviors are tracked within a sliding window of length $W$ using gradient-based statistics.

\paragraph{Gradient-Based Metrics.}
For expert $E_i$ with parameters $\theta_i$, we compute the average gradient norm to measure Expert Optimization Contribution (EOC) over the window:
\begin{equation}
\mathcal{I}_{\mathrm{curr}}^{(i)} = \frac{1}{W} \sum_{k=1}^{W} 
\left\| \frac{\partial \mathcal{L}(\mathcal{D}_k)}{\partial \theta_i} \right\|_2.
\label{eq:I_curr}
\end{equation}

We compute the gradient variance to measure Expert Optimization Instability (EOI) over the window:
\begin{equation}
\mathcal{V}_{\mathrm{curr}}^{(i)} =
\frac{1}{W} \sum_{k=1}^{W} 
\left\| \frac{\partial \mathcal{L}(\mathcal{D}_k)}{\partial \theta_i} \right\|_2^2
-
\left\| \frac{1}{W} \sum_{k=1}^{W} 
\frac{\partial \mathcal{L}(\mathcal{D}_k)}{\partial \theta_i} \right\|_2^2 ,
\label{eq:V_curr}
\end{equation}
where $\mathcal{D}_k$ denotes the mini-batch of training data sampled at the $k$-th training step within the window.

\paragraph{Historical Adaptive Baselines.}
For each expert, we maintain exponential moving averages as historical references:
\begin{equation}
\begin{aligned}
\mathcal{H}_{\mathcal{I}}^{(i)} &= \alpha \, \mathcal{H}_{\mathcal{I}}^{(i)} + (1-\alpha)\, \mathcal{I}_{\mathrm{curr}}^{(i)}, \\
\mathcal{H}_{\mathcal{V}}^{(i)} &= \alpha \, \mathcal{H}_{\mathcal{V}}^{(i)} + (1-\alpha)\, \mathcal{V}_{\mathrm{curr}}^{(i)},
\end{aligned}
\label{eq:H_I_V}
\end{equation}
where $\alpha$ is the momentum coefficient (default $0.9$), and $\mathcal{H}_{\mathcal{I}}^{(i)}$ and $\mathcal{H}_{\mathcal{V}}^{(i)}$ represent long-term reference levels for optimization contribution and instability.

\paragraph{Expert Evolution.}
Within an evolutionary window, SCEE updates the expert pool at fixed intervals.
\vspace{-2pt}
\begin{itemize}[leftmargin=1.2em,itemsep=2pt,topsep=2pt,parsep=0pt,partopsep=0pt]
    \item \textbf{Expert Pruning.}
    An expert is removed if it has low optimization contribution and low activation frequency:
    \vspace{-4pt}
    \begin{equation}
    \mathcal{P}_i =
    \mathbb{I}\left(
    f_i< \gamma_{\mathrm{prune}}
    \right)
    \cdot
    \mathbb{I}\left(
    \frac{\mathcal{I}_{\mathrm{curr}}^{(i)}}{\mathcal{H}_{\mathcal{I}}^{(i)}}  < 1 - \gamma_{\mathrm{prune}}
    \right).
    \label{eq:prune}
    \end{equation}
    \vspace{-8pt}
    
    where $f_i$ denotes the activation frequency of expert $E_i$, and $\gamma_{\mathrm{prune}}$ is the pruning threshold (default $0.05$). Expert $E_i$ is pruned if $\mathcal{P}_i = 1$.

    \item \textbf{Expert Expansion.} 
    A new expert is spawned when a frequently activated expert exhibits unusually high optimization instability:
    \vspace{-4pt}
    \begin{equation}
    f_i \cdot
    \frac{\mathcal{V}_{\mathrm{curr}}^{(i)}}{\mathcal{H}_{\mathcal{V}}^{(i)}}
    >
    \gamma_{\mathrm{expand}}.
    \label{eq:expand}
    \end{equation}
    \vspace{-8pt}
    
    where $\gamma_{\mathrm{expand}}$ is the expansion threshold (default $0.95$). The expanded expert is initialized as 
    $\theta_{\mathrm{new}} = \theta_i + \epsilon$, where $\epsilon \sim \mathcal{N}(0, \sigma^2)$.
\end{itemize}

\begin{table*}[t]
    \centering
    \setlength{\tabcolsep}{1pt}
    \caption{Comparison with state-of-the-art methods on MTIL benchmark (\textbf{Order I}) in terms of ``Transfer'', ``Avg'', and ``Last'' scores (\%). We label the best methods with \textbf{bold} styles. The top block indicates the upper-bound solutions to adapt the CLIP on each task.}
    \label{tab:MTIL-order-I}
    \begin{small}
        \begin{tabular}{
        l
        *{11}{>{\centering\arraybackslash}p{0.8cm}}
        >{\centering\arraybackslash}p{1.4cm}
        }
        \toprule
        \textbf{Method} &
        \rotatebox{45}{Aircraft} &
        \rotatebox{45}{Caltech101} &
        \rotatebox{45}{CIFAR100} &
        \rotatebox{45}{DTD} &
        \rotatebox{45}{EuroSAT} &
        \rotatebox{45}{Flowers} &
        \rotatebox{45}{Food} &
        \rotatebox{45}{MNIST} &
        \rotatebox{45}{OxfordPet} &
        \rotatebox{45}{Cars} &
        \rotatebox{45}{SUN397} &
        Average
        \\
        \midrule
 
        \quad Zero-shot & 24.3 & 88.4 & 68.2 & 44.6 & 54.9 & 71.0 & 88.5 & 59.4 & 89.0 & 64.7 & 65.2 & 65.3 \\
        \quad Fine-tune & 62.0 & 95.1 & 89.6 & 79.5 & 98.9 & 97.5 & 92.7 & 99.6 & 94.7 & 89.6 & 81.8 & 89.2 \\
        \midrule
        \midrule
        
        \textbf{Transfer} & & & & & & & & & & & & \\
        \quad LwF~\cite{li2017learning}     & -- & 74.5 & 56.9 & 39.1 & 51.1 & 52.6 & 72.8 & 60.6 & 75.1 & 30.3 & 55.9 & 58.9 \\
        \quad iCaRL~\cite{rebuffi2017icarl}   & -- & 56.6 & 44.6 & 32.7 & 39.3 & 46.6 & 68.0 & 46.0 & 77.4 & 31.9 & 60.5 & 50.4 \\
        \quad WiSE-FT~\cite{wortsman2022robust} & -- & 73.5 & 55.6 & 35.6 & 41.5 & 47.0 & 68.3 & 53.9 & 69.3 & 26.8 & 51.9 & 52.3 \\
        \quad ZSCL~\cite{zheng2023preventing}    & -- & 86.0 & 67.4 & 45.4 & 50.4 & 69.1 & 87.6 & 61.8 & 86.8 & 60.1 & 66.8 & 68.1 \\
        \quad MoE-Adapters~\cite{yu2024boosting} & -- & 87.9 & 68.2 & 44.4 & 49.9 & \textbf{70.7} & \textbf{88.7} & 59.7 & \textbf{89.1} & 64.5 & 65.5 & 68.9 \\
        \quad GIFT~\cite{wu2025synthetic} & -- & \textbf{88.5} & \textbf{69.8} & \textbf{46.0} & 49.4 & 68.5 & 87.1 & \textbf{69.9} & 88.9 & 57.7 & \textbf{67.7} & \textbf{69.3} \\
        \quad LEBA~\cite{Gu2025gulearn} & -- & \textbf{88.5} & 68.3 & 44.8 & 49.4 & 70.2 & 88.6 & 60.9 & \textbf{89.1} & \textbf{64.8} & 64.2 & 69.2 \\
        \rowcolor{lightblue!80}
        \quad \textbf{Ours} & -- & \textbf{88.5} & 68.2 & 44.7 & \textbf{55.0} & 69.5 & 88.5 & 58.9 & \textbf{89.1} & 64.7 & 65.9 & \textbf{69.3} \\
        \midrule

        \textbf{Avg.} & & & & & & & & & & & & \\
        \quad LwF~\cite{li2017learning}      & 36.3 & 86.9 & 72.0 & 59.0 & 73.7 & 60.0 & 73.6 & 74.8 & 80.0 & 37.3 & 58.1 & 64.7 \\
        \quad iCaRL~\cite{rebuffi2017icarl}   & 35.5 & 89.2 & 72.2 & 60.6 & 68.8 & 70.0 & 78.2 & 62.3 & 81.8 & 41.2 & 62.5 & 65.7 \\
        \quad WiSE-FT~\cite{ding2022don} & 26.7 & 86.5 & 64.3 & 57.1 & 65.7 & 58.7 & 71.1 & 70.5 & 75.8 & 36.9 & 54.6 & 60.7 \\
        \quad ZSCL~\cite{zheng2023preventing}     & 45.1 & 92.0 & 80.1 & 64.3 & 79.5 & 81.6 & 89.6 & 75.2 & 88.9 & 64.7 & 68.0 & 75.4 \\
        \quad MoE-Adapters~\cite{yu2024boosting} & 50.2 & 91.9 & 83.1 & 69.4 & 78.9 & 84.0 & 89.1 & 73.7 & 89.3 & 67.7 & 66.9 & 76.7 \\
        \quad GIFT~\cite{wu2025synthetic} & 51.9 & 93.9 & 81.4 & 67.7 & 80.3 & 82.8 & 89.3 & \textbf{80.6} & \textbf{90.3} & 63.1 & \textbf{68.9} & 77.3 \\
        \quad LEBA~\cite{Gu2025gulearn} & 53.9 & 94.9 & 83.8 & \textbf{70.8} & 79.8 & \textbf{85.1} & 89.1 & 74.8 & 89.3 & \textbf{69.2} & 65.8 & 77.9 \\
        \rowcolor{lightblue!80}
        \quad \textbf{Ours} & \textbf{61.6} & \textbf{96.4} & \textbf{85.1} & 69.7 & \textbf{82.9} & 84.3 & \textbf{90.4} & 73.7 & 90.0 & 69.1 & 67.4 & \textbf{79.1} \\
        \midrule

        \textbf{Last} & & & & & & & & & & & & \\
        \quad LwF~\cite{li2017learning}      & 26.3 & 87.5 & 71.9 & 66.6 & 79.9 & 66.9 & 83.8 & \textbf{99.6} & 92.1 & 66.1 & 80.4 & 74.6 \\
        \quad iCaRL~\cite{rebuffi2017icarl}   & 35.8 & 93.0 & 77.0 & 70.2 & 83.3 & 88.5 & 90.4 & 86.7 & 93.2 & 81.2 & 81.9 & 80.1 \\
        \quad WiSE-FT~\cite{ding2022don} & 27.2 & 90.8 & 68.0 & 68.9 & 86.9 & 74.0 & 87.6 & \textbf{99.6} & 92.6 & 77.8 & 81.3 & 77.7 \\
        \quad ZSCL~\cite{zheng2023preventing}   & 40.6 & 92.2 & 81.3 & 70.5 & 94.8 & 90.5 & 91.9 & 98.7 & 93.9 & 85.3 & 80.2 & 83.6 \\
         \quad MoE-Adapters~\cite{yu2024boosting} & 49.8 & 92.2 & 86.1 & 78.1 & 95.7 & 94.3 & 89.5 & 98.1 & 89.9 & 81.6 & 80.0 & 85.0 \\
        \quad GIFT~\cite{wu2025synthetic} & 47.9 & 95.6 & 82.8 & 75.1 & 97.3 & 94.2 & 91.7 & 99.2 & \textbf{94.2} & 87.0 & 80.9 & 86.0 \\
        \quad LEBA~\cite{Gu2025gulearn} & 55.1 & 95.2 & 87.4 & 78.8 & 97.2 & \textbf{97.3} & 89.5 & 99.1 & 89.6 & \textbf{88.8} & 82.4 & 87.3 \\
        \rowcolor{lightblue!80}
        \quad \textbf{Ours} & \textbf{61.7} & \textbf{97.2} & \textbf{88.8} & \textbf{79.1} & \textbf{98.9} & 96.6 & \textbf{92.6} & \textbf{99.6} & 92.6 & \textbf{88.8} & \textbf{82.9} & \textbf{89.0}\\
        \midrule

    \end{tabular}
    \end{small}
\end{table*}
\subsection{Prototype-Guided Expert Selection}
The proposed DIMoE-Adapters require selecting task-specific expert routes at inference. Therefore, we introduce Prototype-Guided Expert Selection (PGES), a training-free mechanism that infers the target task based on feature distributions, thereby enabling reliable expert selection.

\paragraph{Prototype Construction.}
For each task $\tau^{(t)}$, the features $\mathcal{X}^{(t)}$ are partitioned into $K$ clusters.
Each cluster is modeled as a multivariate Gaussian component, yielding the task-specific prototypes:
\begin{equation}
\begin{gathered}
\boldsymbol{\mu}_{t,k}
= \frac{1}{|\mathcal{C}_k|}
\sum_{\mathbf{x} \in \mathcal{C}_k} \mathbf{x},\\
\boldsymbol{\Sigma}_{t,k}
= \frac{1}{|\mathcal{C}_k|-1}
\sum_{\mathbf{x} \in \mathcal{C}_k}
(\mathbf{x} - \boldsymbol{\mu}_{t,k})
(\mathbf{x} - \boldsymbol{\mu}_{t,k})^\top,
\label{eq:task_Prototype}
\end{gathered}
\end{equation}
where $\mathcal{C}_k$ denotes the set of features assigned to the $k$-th cluster, $|\mathcal{C}_k|$ is the number of samples in this cluster, $\boldsymbol{\mu}_{t,k}$ is the mean vector of cluster $k$ for task $\tau^{(t)}$, and $\boldsymbol{\Sigma}_{t,k}$ is the corresponding covariance matrix.

\paragraph{Inference.}
Given a test sample $\mathbf{x}$, PGES computes its log-likelihood under each task-specific Gaussian mixture.
The task confidence score is defined as:
\begin{equation}
\begin{gathered}
    \mathcal{S}_t(\mathbf{x}) = \max_{k} \left( -\frac{1}{2} d_{t,k}^2(\mathbf{x}) - \frac{1}{2} \ln \lvert \boldsymbol{\Sigma}_{t,k} \rvert \right), \\
    \text{with} \quad d_{t,k}^2(\mathbf{x}) = (\mathbf{x} - \boldsymbol{\mu}_{t,k})^\top \boldsymbol{\Sigma}_{t,k}^{-1} (\mathbf{x} - \boldsymbol{\mu}_{t,k}),
\label{eq:Inference}
\end{gathered}
\end{equation}
where $d_{t,k}^2(\mathbf{x})$ represents the squared Mahalanobis distance. The predicted task is $\hat{t} = \arg\max_t \mathcal{S}_t(\mathbf{x})$.
If the maximum score is below a threshold $\delta$, the sample is routed to the frozen CLIP model.
Otherwise, the corresponding task-specific router $\mathcal{R}^{(\hat{t})}$ is activated.

\begin{table*}[t]
    \centering
    \setlength{\tabcolsep}{1pt}
    \caption{Comparison with state-of-the-art methods on few-shot MTIL benchmark (\textbf{Order I}) in terms of ``Transfer'', ``Avg.'', and ``Last'' scores (\%). We label the best methods with \textbf{bold} styles. The top block indicates the upper-bound solutions to adapt the CLIP on each task.}
    \label{tab:FS-MTIL-order-I}
    \begin{small}
        \begin{tabular}{
        l
        *{11}{>{\centering\arraybackslash}p{0.8cm}}
        >{\centering\arraybackslash}p{1.4cm}
        }
    \toprule
        \textbf{Method} &
        \rotatebox{45}{Aircraft} &
        \rotatebox{45}{Caltech101} &
        \rotatebox{45}{CIFAR100} &
        \rotatebox{45}{DTD} &
        \rotatebox{45}{EuroSAT} &
        \rotatebox{45}{Flowers} &
        \rotatebox{45}{Food} &
        \rotatebox{45}{MNIST} &
        \rotatebox{45}{OxfordPet} &
        \rotatebox{45}{Cars} &
        \rotatebox{45}{SUN397} &
        Average
        \\
        \midrule
        
        \quad Zero-shot & 24.3 & 88.4 & 68.2 & 44.6 & 54.9 & 71.0 & 88.5 & 59.4 & 89.0 & 64.7 & 65.2 & 65.3 \\
        \quad $5$-shot Full Fine-tune & 30.6 & 93.5 & 76.8 & 65.1 & 91.7 & 92.9 & 83.3 & 96.6 & 84.9 & 65.4 & 71.3 & 77.5 \\
        \midrule
        \midrule
        
        \textbf{Transfer} & & & & & & & & & & & & \\
        \quad LwF~\cite{li2017learning} & -- & 72.1 & 49.2 & 35.9 & 44.5 & 41.1 & 66.6 & 50.5 & 69.0 & 19.0 & 51.7 & 50.0 \\
        \quad LwF-VR~\cite{ding2022don} & -- & 82.2 & 62.5 & 40.1 & 40.1 & 56.3 & 80.0 & 60.9 & 77.6 & 40.5 & 60.8 & 60.1 \\
        \quad WiSE-FT~\cite{wortsman2022robust} & -- & 77.6 & 60.0 & 41.3 & 39.4 & 53.0 & 76.6 & 58.1 & 75.5 & 37.3 & 58.2 & 57.7 \\
        \quad ZSCL~\cite{zheng2023preventing} & -- & 84.0 & 68.1 & \textbf{44.8} & 46.8 & 63.6 & 84.9 & 61.4 & 81.4 & 55.5 & 62.2 & 65.3 \\
        \quad MoE-Adapters~\cite{yu2024boosting} & -- & 87.9 & \textbf{68.2} & 44.1 & 48.1 & 64.7 & \textbf{88.8} & \textbf{69.0} & \textbf{89.1} & 64.5 & 65.1 & 68.9 \\
        \rowcolor{lightblue!80}
        \quad \textbf{Ours} & -- & \textbf{88.5} & \textbf{68.2} & 44.7 & \textbf{55.0} & \textbf{71.0} & 88.5 & 58.9 & \textbf{89.1} & \textbf{64.7} & \textbf{65.9} & \textbf{69.5} \\
        \midrule

        \textbf{Avg.} & & & & & & & & & & & & \\
        \quad LwF~\cite{li2017learning} & 23.5 & 77.4 & 43.5 & 41.7 & 43.5 & 52.2 & 54.6 & 63.4 & 68.0 & 21.3 & 52.6 & 49.2 \\
        \quad LwF-VR~\cite{ding2022don} & 24.9 & 89.1 & 64.2 & 53.4 & 54.3 & 70.8 & 79.2 & 66.5 & 79.2 & 44.1 & 61.6 & 62.5 \\
        \quad WiSE-FT~\cite{wortsman2022robust} & 32.0 & 87.7 & 61.0 & 55.8 & 68.1 & 69.3 & 76.8 & 71.5 & 77.6 & 42.0 & 59.3 & 63.7 \\
        \quad ZSCL~\cite{zheng2023preventing} & 28.2 & 88.6 & 66.5 & 53.5 & 56.3 & 73.4 & 83.1 & 56.4 & 82.4 & 57.5 & 62.9 & 64.4 \\
        \quad MoE-Adapters~\cite{yu2024boosting} & 30.0 & 89.6 & 73.9 & 58.7 & 69.3 & 79.3 & 88.1 & \textbf{76.5} & 89.1 & 65.3 & 65.8 & 71.4 \\
        \rowcolor{lightblue!80}
        \quad \textbf{Ours} & \textbf{35.3} & \textbf{92.5} & \textbf{74.7} & \textbf{59.3} & \textbf{74.6} & \textbf{81.0} & \textbf{88.4} & 70.4 & \textbf{89.8} & \textbf{65.4} & \textbf{66.5} & \textbf{72.5} \\
        \midrule

        \textbf{Last} & & & & & & & & & & & & \\
        \quad LwF~\cite{li2017learning} & 22.1 & 58.2 & 17.9 & 32.1 & 28.1 & 66.7 & 46.0 & 84.3 & 64.1 & 31.5 & 60.1 & 46.5 \\
        \quad LwF-VR~\cite{ding2022don} & 22.9 & 89.8 & 59.3 & 57.1 & 57.6 & 79.2 & 78.3 & 77.7 & 83.6 & 60.1 & 69.8 & 66.9 \\
        \quad WiSE-FT~\cite{wortsman2022robust} & 30.8 & 88.9 & 59.6 & 60.3 & 80.9 & 81.7 & 77.1 & \textbf{94.9} & 83.2 & 62.8 & 70.0 & 71.9 \\ 
        \quad ZSCL~\cite{zheng2023preventing} & 26.8 & 88.5 & 63.7 & 55.7 & 60.2 & 82.1 & 82.6 & 58.6 & 85.9 & 66.7 & 70.4 & 67.4 \\
        \quad MoE-Adapters~\cite{yu2024boosting} & 30.1 & 89.3 & 74.9 & 64.0 & 82.3 & \textbf{89.4} & 87.1 & 89.0 & 89.1 & \textbf{69.5} & \textbf{72.5} & 76.1 \\
        \rowcolor{lightblue!80}
        \quad \textbf{Ours} & \textbf{35.3} & \textbf{92.9} & \textbf{76.1} & \textbf{64.7} & \textbf{86.0} & 89.3 & \textbf{88.2} & 90.8 & \textbf{91.6} & 68.7 & 72.4 & \textbf{77.8} \\
        \midrule

    \end{tabular}
    \end{small}
\end{table*}
\section{Experiments}
\subsection{Experimental Setting}
\paragraph{Datasets.}
We evaluate our method under three continual learning settings: Multi-domain Task-Incremental Learning (MTIL), Class-Incremental Learning (CIL), and Domain-Incremental Learning (DIL). MTIL is a challenging benchmark for vision--language continual learning, comprising 11 datasets across diverse domains with 1,201 classes. We follow the two-order training protocol defined in the original benchmark.
We further adopt a few-shot MTIL setting with 5 samples per class to evaluate performance under limited data. Results for CIL and DIL are provided in the supplementary material (see \autoref{sec:supp-cil-results} and \autoref{sec:supp-dil-results}).

\paragraph{Metrics.}
For MTIL and DIL, we adopt the metrics from~\cite{zheng2023preventing}, including ``Transfer'', ``Avg.'', and ``Last''. ``Transfer'' measures  zero-shot generalization performance on unseen tasks, while ``Last'' reflects retention of previously learned knowledge. ``Avg.'' summarizes overall performance across tasks.
For CIL, following~\cite{douillard2022dytox}, we report the average accuracy over all steps(``Average''), and the accuracy on the final steps (``Last'').

\paragraph{Implementation Details.}
Following~\cite{zheng2023preventing}, we adopt CLIP with a ViT-B/16 backbone.
Each expert is implemented as a LoRA~\cite{hu2022lora} module, and the router is a single-layer MLP with Top-$p$ gating.
SCEE dynamically updates the expert pool during training.
We use the AdamW optimizer~\cite{loshchilov2017decoupled} with label smoothing~\cite{muller2019does}, and all experiments are conducted on a single NVIDIA A6000 GPU.
Additional implementation details are provided in the supplementary material (see \autoref{sec:supp-implementation-details}).

\subsection{Comparison with State-of-the-art Methods}
\paragraph{Multi-domain Task-Incremental Learning.}
\autoref{tab:MTIL-order-I} compares our method with traditional continual learning approaches and methods specifically designed for the MTIL setting under the Order-I protocol~\cite{zheng2023preventing}.
In this setting, tasks are learned sequentially, following a fixed order.
Additional results under Order-II are provided in the supplementary material (\autoref{sec:supp-MTIL-results}).

The upper block of \autoref{tab:MTIL-order-I} reports the results obtained via zero-shot inference and full fine-tuning.
The ``Zero-shot'' results reflect CLIP's inherent zero-shot transferability, while ``Fine-tune'' represents the best performance achievable by fully fine-tuning CLIP on each individual task.

As shown in \autoref{tab:MTIL-order-I}, our method achieves the best overall performance across most datasets.
Compared with MoE-Adapters~\cite{yu2024boosting}, our method improves the ``Transfer'', ``Avg.'', and ``Last'' metrics by 0.4\%, 2.4\%, and 4.0\%, respectively.
Moreover, our method consistently outperforms existing state-of-the-art approaches on MTIL Order-I, achieving the highest overall scores across all three evaluation metrics.
These results indicate that our approach achieves a favorable balance between plasticity and stability,
by dynamically allocating model capacity across tasks while mitigating interference between domains.

\paragraph{Few-shot Multi-domain Task-Incremental Learning.}
We further evaluate our method under the few-shot MTIL setting, where only a limited number of training samples are available for each task.
All experiments follow the same Order-I protocol and evaluation metrics as in the full-shot setting.
Under the 5-shot setting, the comparison results are summarized in \autoref{tab:FS-MTIL-order-I}, with additional Order-II results reported in the supplementary material(\autoref{sec:supp-MTIL-results}).

As shown in \autoref{tab:FS-MTIL-order-I}, our method achieves the best overall performance across most datasets.
When compared with MoE-Adapters, our method improves the ``Transfer'', ``Avg.'', and ``Last'' metrics by 0.6\%, 1.1\%, and 1.7\%, respectively.
These results indicate that our method more effectively mitigates forgetting in the few-shot MTIL setting.

\begin{wraptable}{r}{0.6\textwidth}
    \vspace{-10pt}
    \centering
    \caption{
    Comparison of computational cost during training.
    }
    \label{tab:efficiency}

    \setlength{\tabcolsep}{1.5pt}
    \begin{small}
    \begin{tabular}{
    >{\centering\arraybackslash}p{2.1cm}
    >{\centering\arraybackslash}p{1.9cm}
    >{\centering\arraybackslash}p{1.9cm}
    >{\centering\arraybackslash}p{1.9cm}
    }
    \toprule
    Method 
    & Train Params $\downarrow$  
    & GPU $\downarrow$  
    & Train Times $\downarrow$  \\
    \midrule
    LWF 
    & 149.6M 
    & 32172MiB
    & 1.54s/it \\
    ZSCL 
    & 149.6M 
    & 26290MiB 
    & 3.94s/it \\
    MoE-Adapters 
    & 59.8M 
    & 22358MiB 
    & 1.58s/it  \\
    \midrule
    \rowcolor{lightblue!80}
    Ours 
    & \textbf{39.5M} 
    & \textbf{16352MiB} 
    & \textbf{1.24s/it} \\
    \bottomrule
    \end{tabular}
    \vspace{-10pt}
    \end{small}
\end{wraptable}

\paragraph{Computational Cost.}
\autoref{tab:efficiency} compares the computational cost of our method with other approaches.
Compared with MoE-Adapters~\cite{yu2024boosting}, our approach achieves reductions of 33.9\% in trainable parameters, 26.9\% in GPU memory usage, and 21.5\% in training time per iteration.
These results demonstrate that DIMoE-Adapters achieve strong performance in MTIL while reducing training costs.

\subsection{Ablation Study}

\begin{wraptable}{r}{0.6\textwidth}
\vspace{-10pt}
\centering
\caption{Ablation studies on different modules.}
\label{tab:analysis-dimoe}
\setlength{\tabcolsep}{1.5pt}
\begin{small}
\begin{tabular}{l
>{\centering\arraybackslash}p{0.9cm}
>{\centering\arraybackslash}p{0.7cm}
>{\centering\arraybackslash}p{0.9cm}
>{\centering\arraybackslash}p{0.7cm}
>{\centering\arraybackslash}p{0.9cm}
>{\centering\arraybackslash}p{0.7cm}}
    \toprule
    Method 
    & Transfer & $\Delta$ 
    & Avg. & $\Delta$ 
    & Last & $\Delta$ \\
    \midrule
    CLIP Zero-shot
    & 69.4 & \textcolor{mydarkred}{+0.1}
    & 65.3 & \textcolor{mydarkblue}{-13.8}
    & 65.3 & \textcolor{mydarkblue}{-23.7} \\
    
    \phantom{+}+Adapter
    & 49.0 & \textcolor{mydarkblue}{-20.3}
    & 58.9 & \textcolor{mydarkblue}{-20.2}
    & 72.5 & \textcolor{mydarkblue}{-16.5} \\
    
    \phantom{+}+Single-R
    & 54.9 & \textcolor{mydarkblue}{-14.4}
    & 61.3 & \textcolor{mydarkblue}{-17.8}
    & 68.5 & \textcolor{mydarkblue}{-20.5} \\
    
    \phantom{+}+Single-R + PGES
    & 69.1 & \textcolor{mydarkblue}{-0.2}
    & 68.1 & \textcolor{mydarkblue}{-11.0}
    & 68.0 & \textcolor{mydarkblue}{-21.0} \\
    
    \phantom{+}+Task-R + PGES
    & 69.0 & \textcolor{mydarkblue}{-0.3}
    & 76.9 & \textcolor{mydarkblue}{-2.2}
    & 85.6 & \textcolor{mydarkblue}{-3.4} \\
    \midrule
    \rowcolor{lightblue!80}
    \phantom{+}\textbf{Ours (Full)}
    & \textbf{69.3} & 0.0
    & \textbf{79.1} & 0.0
    & \textbf{89.0} & 0.0 \\
    \bottomrule
\end{tabular}
\vspace{-10pt}
\end{small}
\end{wraptable}

\paragraph{Analysis of DIMoE-Adapters}
We analyze design choices of DIMoE-Adapters with
Top-$p$ routing and task-aware load balancing, as summarized in
\autoref{tab:analysis-dimoe}.
The Single-R configuration employs a shared router whose output dimension expands as new tasks arrive, whereas Task-R utilizes task-specific routers that are introduced incrementally.

Compared with Single-R, adding PGES substantially improves the ``Transfer'' score,
bringing it close to the CLIP zero-shot baseline.
This indicates that PGES routes test samples to either DIMoE-Adapters or the frozen CLIP, thereby preserving zero-shot generalization.

Replacing Single-R with Task-R further improves the ``Avg.'' and ``Last'' metrics,
suggesting reduced cross-task interference and better retention of task-specific knowledge.

Finally, combining Task-R, PGES, and SCEE achieves the best performance across all metrics.
These results demonstrate the effectiveness of SCEE in regulating expert evolution based on expert behavior.

\begin{wraptable}{r}{0.6\textwidth}
\vspace{-10pt}
\centering
\caption{Ablation studies on SCEE components.}
\label{tab:analysis-scee-dimoe}
\setlength{\tabcolsep}{1.5pt}
\begin{small}
\begin{tabular}{
cccc|
>{\centering\arraybackslash}p{0.9cm}
>{\centering\arraybackslash}p{0.7cm}
>{\centering\arraybackslash}p{0.9cm}
>{\centering\arraybackslash}p{0.7cm}
>{\centering\arraybackslash}p{0.9cm}
>{\centering\arraybackslash}p{0.7cm}
}
\toprule
\multicolumn{4}{c|}{Configuration} & \multicolumn{6}{c}{Metrics} \\
\cmidrule(lr){1-4} \cmidrule(lr){5-10}
Add & Prune & LB & Router
& Transfer & $\Delta$
& Avg.& $\Delta$
& Last & $\Delta$ \\
\midrule

\xmark & \xmark & \checkmark & Top-\emph{p}
& 69.0  & \textcolor{mydarkblue}{-0.3}
& 76.9 & \textcolor{mydarkblue}{-2.2}
& 85.6 & \textcolor{mydarkblue}{-3.4} \\

\checkmark & \xmark & \checkmark & Top-\emph{p}
& 68.9 & \textcolor{mydarkblue}{-0.4}
& 77.6 & \textcolor{mydarkblue}{-1.5}
& 87.3 & \textcolor{mydarkblue}{-1.7} \\

\xmark & \checkmark & \checkmark & Top-\emph{p}
& 69.4 & \textcolor{mydarkred}{+0.1}
& 77.2 & \textcolor{mydarkblue}{-1.9}
& 86.6 & \textcolor{mydarkblue}{-2.4} \\

\midrule

\checkmark & \checkmark & \textcolor{gray}{\xmark} & Top-\emph{p}
& 69.1 & \textcolor{mydarkblue}{-0.2}
& 78.7 & \textcolor{mydarkblue}{-0.4}
& 88.0 & \textcolor{mydarkblue}{-1.0} \\

\midrule

\checkmark & \checkmark & \checkmark & \textcolor{gray}{Top-2}
& 69.2 & \textcolor{mydarkblue}{-0.1}
& 78.4 & \textcolor{mydarkblue}{-0.7}
& 87.5 & \textcolor{mydarkblue}{-1.5} \\

\midrule

\rowcolor{lightblue!80}
\checkmark & \checkmark & \checkmark & Top-\emph{p}
& \textbf{69.3} & 0.0
& \textbf{79.1} & 0.0
& \textbf{89.0} & 0.0 \\

\bottomrule
\end{tabular}
\vspace{-10pt}
\end{small}
\end{wraptable}

\paragraph{Analysis of SCEE for DIMoE-Adapters.}
We conduct ablation studies on the key components of SCEE,
with results summarized in \autoref{tab:analysis-scee-dimoe}.
We analyze the roles of expert expansion, expert pruning, task-aware load balancing, and routing strategies.

When only expert expansion is enabled, the model exhibits improved adaptation to new tasks.
However, without pruning, redundant experts accumulate as training progresses, which dilutes routing selectivity and limits gains on the ``Avg.'' and ``Last'' metrics.
In contrast, enabling expert pruning alone improves model stability by reducing interference, but the lack of additional capacity restricts adaptation to new tasks, leading to lower overall performance.

When both expert expansion and pruning are enabled, removing task-aware load balancing or replacing Top-$p$ routing with fixed Top-2 routing results in degraded long-term performance, particularly on the ``Last'' metric.
This indicates that effective expert utilization requires both balanced expert participation and adaptive routing decisions.

The full configuration achieves the best performance across ``Transfer'', ``Avg.'', and ``Last'' metrics.
By jointly introducing new experts when needed and removing redundant or ineffective ones, while regulating expert participation through load balancing and Top-$p$ routing, SCEE maintains sufficient capacity for new tasks and mitigates interference among experts.
This regulation enables stable and effective continual learning in DIMoE-Adapters.

\begin{wrapfigure}{r}{0.5\columnwidth}
\vspace{-10pt}
\centering
\includegraphics[width=1.0\linewidth]{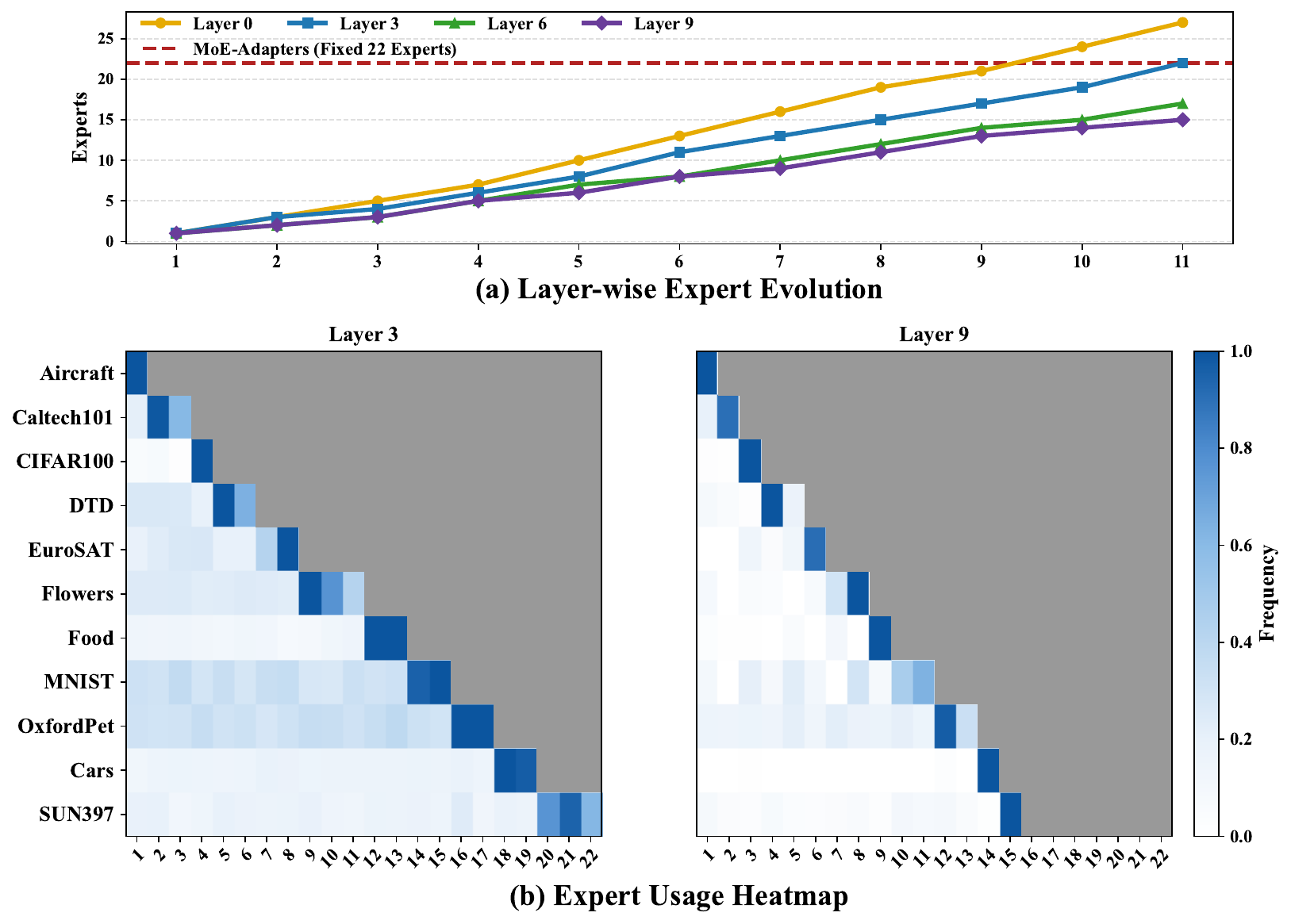}
\caption{
Layer-wise behavior of dynamic expert evolution.
(a) Number of experts across layers.
(b) Expert usage heatmaps.
}
\label{fig:layer3_layer9_expert_usage_heatmap_compact}
\vspace{-20pt}
\end{wrapfigure}

\paragraph{Layer-wise Behavior of Dynamic Expert Evolution.}
\autoref{fig:layer3_layer9_expert_usage_heatmap_compact} reveals a clear layer-wise behavior in expert evolution. As shown in \autoref{fig:layer3_layer9_expert_usage_heatmap_compact}(a), in contrast to the fixed expert allocation adopted by MoE-Adapters, DIMoE-Adapters progressively accumulate more experts in shallow layers, while maintaining a more compact expert set in deeper layers.
This behavior indicates a depth-dependent capacity allocation strategy induced by SCEE, where model capacity is adaptively redistributed across layers.

\autoref{fig:layer3_layer9_expert_usage_heatmap_compact}(b) further visualizes the corresponding expert usage patterns through expert usage heatmaps.
In shallow layers, the representations mainly capture low- and mid-level visual statistics that are shared across domains~\cite{zeiler2014visualizing,yosinski2014transferable}.
When new tasks are introduced, variations in textures, backgrounds, and local patterns lead to more distributed routing decisions across experts.
As a result, a larger set of experts is activated, and SCEE introduces new experts to model the newly observed distributions. In contrast, deeper layers encode higher-level semantic information~\cite{raghu2019transfusion}.
Routing decisions in these layers are typically concentrated on a small number of experts, indicating more stable and task-specific semantic responses.
Accordingly, SCEE triggers expert expansion less frequently at greater depths and instead prunes experts with low contribution, maintaining a compact expert set focused on semantic discrimination. We provide more comprehensive visualizations of expert evolution in the supplementary material(\autoref{sec:additional-experimetal-results}).

\begin{wrapfigure}{r}{0.5\columnwidth}
\vspace{-10pt}
\centering
\includegraphics[width=\linewidth]{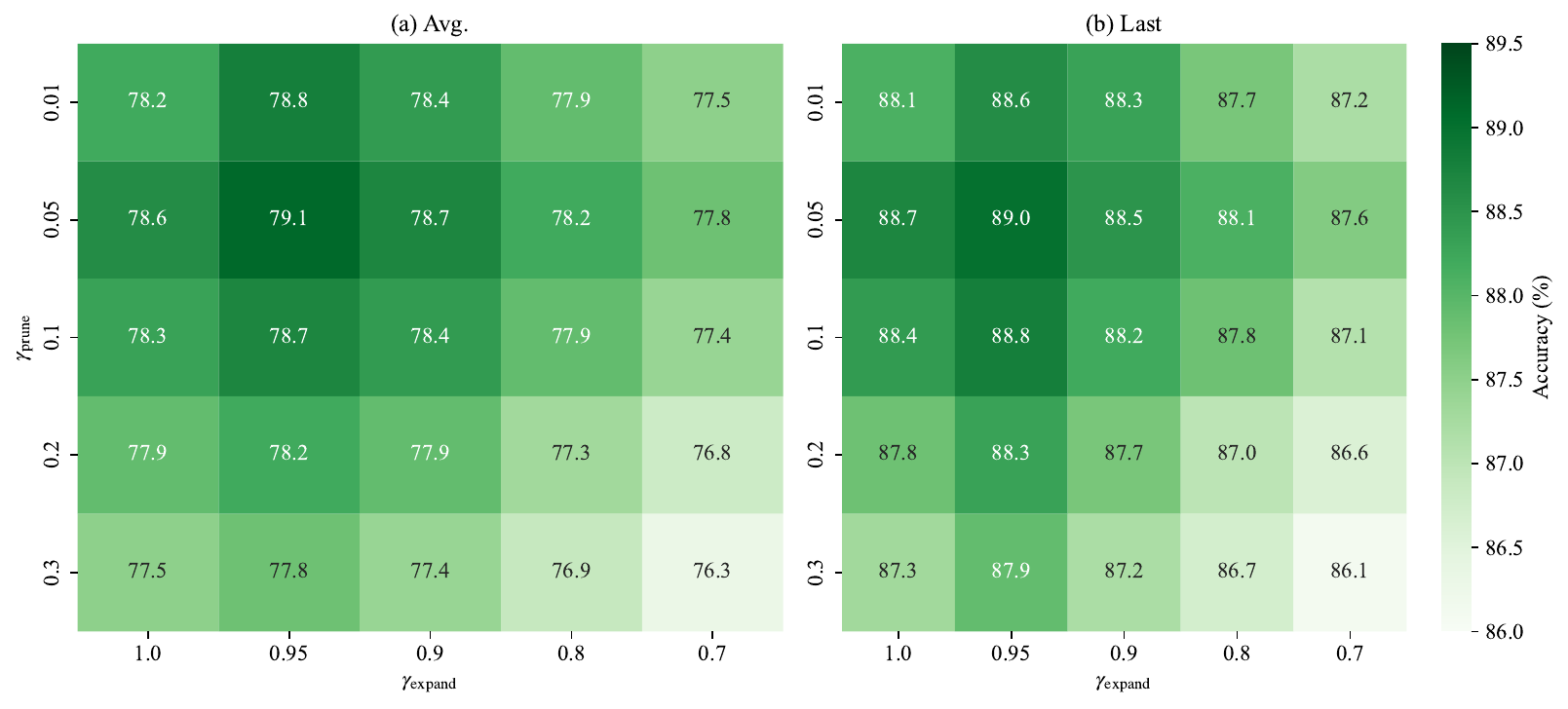}
\vspace{-8pt}
\caption{
Joint sensitivity analysis of $\gamma_{\text{expand}}$ and $\gamma_{\text{prune}}$.
(a) Avg. accuracy. (b) Last accuracy.
}
\label{fig:sensitivity_avg_last}
\vspace{-10pt}
\end{wrapfigure}

\paragraph{Hyperparameter Sensitivity Analysis.}
We analyze the joint sensitivity of DIMoE-Adapters to  $\gamma_{\text{expand}}$ and $\gamma_{\text{prune}}$.
As shown in \autoref{fig:sensitivity_avg_last}.

The best performance is achieved at $\gamma_{\text{expand}}=0.95$ and $\gamma_{\text{prune}}=0.05$, indicating a balanced trade-off between expert expansion and pruning.
When $\gamma_{\text{expand}}$ is too small, frequent expansion introduces redundant experts and increases interference.
Conversely, overly large values delay expansion and limit the model's adaptability to new tasks.
Similarly, a large $\gamma_{\text{prune}}$ aggressively removes experts, leading to underfitting, while overly small values retain redundant experts and reduce efficiency. We present additional sensitivity analyses and more comprehensive experimental results in the supplementary material(\autoref{sec:additional-experimetal-results}).

\section{Conclusion}
In this paper, we propose DIMoE-Adapters, a dynamic mixture-of-experts framework for continual learning of vision--language models under significant domain shifts.
By introducing a Dynamic Expert Evolution paradigm, DIMoE-Adapters balance stability and plasticity in MTIL while preserving zero-shot generalization.
Through Self-Calibrated Expert Evolution (SCEE) and Prototype-Guided Expert Selection (PGES), the framework dynamically adapts model capacity during training and ensures reliable expert utilization at inference.
Extensive experiments on MTIL and standard continual learning benchmarks demonstrate consistent improvements over existing state-of-the-art methods, and layer-wise analyses further reveal meaningful depth-dependent expert evolution patterns.

\section*{Impact Statement}

This work proposes DIMoE-Adapters, a dynamic continual learning framework for CLIP-based VLMs.
All training and evaluation are conducted on publicly available benchmark datasets under standard experimental settings that are consistent with prior continual learning methods.
The proposed approach focuses on algorithmic improvements and does not involve new data collection or real-world deployment, and therefore does not introduce additional privacy, security, or ethical concerns beyond those commonly considered in existing continual learning research.
However, our method still has some limitations: it is relatively more complex than standard adapter-based methods, since it involves task identification, dynamic expert expansion, and expert pruning, and it also introduces several hyperparameters, such as the expansion threshold, pruning threshold, and confidence threshold.
\bibliographystyle{unsrtnat}
\bibliography{example_paper}

\clearpage
\appendix

\section{Additional Implementation Details}
\label{sec:supp-implementation-details}
\paragraph{Hyperparameter Settings.}
For the MTIL benchmark, we use a batch size of 128 during training and 256 during evaluation.
All models are optimized using AdamW~\cite{loshchilov2017decoupled} with $\beta_1 = 0.9$, $\beta_2 = 0.999$, and a weight decay of 0.
Label smoothing~\cite{muller2019does} is employed to improve training stability and generalization, with the smoothing factor set to 0.2. 
The expert expansion threshold is set to $\gamma_{\text{expand}} = 0.95$, while the pruning threshold is $\gamma_{\text{prune}} = 0.05$.
For the Historical Adaptive Baseline, an exponential moving average~\cite{polyak1992acceleration} with a momentum coefficient $\alpha = 0.9$ is maintained.
The task-aware load balancing coefficient for newly introduced experts is set to $\beta = 0.6$.

\paragraph{Expert Architecture.}
Each expert is implemented as a low-rank adapter inserted into the feed-forward layers of the pretrained vision--language model.
A LoRA-based expert consists of two learnable linear projections: a down-projection matrix $\mathbf{W}_{\text{down}} \in \mathbb{R}^{d \times r}$ and an up-projection matrix $\mathbf{W}_{\text{up}} \in \mathbb{R}^{r \times d}$, where $r \ll d$ denotes the rank of the low-rank decomposition.

Given an input feature sequence $\mathbf{x} \in \mathbb{R}^{L \times d}$, the output $\mathbf{y} \in \mathbb{R}^{L \times d}$ of the expert is computed as:
\begin{equation}
\mathbf{E}(\mathbf{x}) = \mathbf{x} \mathbf{W}_{\text{down}} \mathbf{W}_{\text{up}},
\end{equation}
where $L$, $d$, and $r$ denote the sequence length, original feature dimension, and low-rank dimension, respectively.

\section{Results on CIL Benchmarks}
\label{sec:supp-cil-results}
\paragraph{Benchmark Description.}
We further evaluate our method under the class-incremental learning (CIL) setting to verify its effectiveness in single-domain continual learning scenarios.
Following the protocols of prior work~\cite{rebuffi2017icarl,buzzega2020dark}, experiments are conducted on CIFAR100~\cite{krizhevsky2009learning} and Tiny-ImageNet~\cite{yan2021dynamically}.
For CIFAR100, the 100 classes are incrementally introduced in steps of 10, 20, and 50 classes.
Similarly, for Tiny-ImageNet, step sizes of 5, 10, and 20 classes are adopted to evaluate robustness under different incremental granularities.
At each incremental step, the model is trained only on the current task data and evaluated on all previously seen classes.
Following standard practice~\cite{rebuffi2017icarl}, we report the average accuracy over all incremental steps (``Avg.'') and the final accuracy after learning all tasks (``Last'').

Different from the MTIL setting, the task identity of each input image is unknown during inference in CIL.
Therefore, task-specific routing cannot be directly applied.
To adapt our framework to this setting, our MoE-Adapters use a single shared router and initialize two experts to handle all class subsets.
During training, the proposed SCEE strategy dynamically adjusts the number of experts according to the evolving data distribution, enabling the model to expand its capacity when new class distributions are introduced while avoiding unnecessary expert redundancy.

\paragraph{Compared Methods.}
We compare our approach with representative state-of-the-art CIL methods, including knowledge distillation-based methods such as LwF~\cite{li2017learning} and LwF-VR~\cite{dhar2019learning}, exemplar-based methods such as iCaRL~\cite{rebuffi2017icarl}, and recent CLIP-based continual learning approaches including ZSCL~\cite{zheng2023preventing} and MoE-Adapters~\cite{yu2024boosting}.
In addition, we report the performance of CLIP zero-shot inference without adaptation, denoted as CLIP Zero-shot, and continual fine-tuning without any forgetting mitigation strategy, denoted as Continual-FT.
For a fair comparison, all CLIP-based methods, including ours, are implemented with the CLIP ViT-B/16 backbone~\cite{radford2021learning} and follow the same training protocol.
The comparison results on CIFAR100 and Tiny-ImageNet are reported in \autoref{tab:cil_results}.

\paragraph{Result Analysis.}
As shown in \autoref{tab:cil_results}, our method consistently achieves the best performance across all CIL settings on both CIFAR100 and Tiny-ImageNet.
Compared with prior CLIP-based methods, our approach obtains clear improvements in both Avg.\ and Last metrics, demonstrating its ability to mitigate catastrophic forgetting while maintaining strong adaptability to newly introduced classes.
The advantage becomes more pronounced when the incremental step size decreases, where the task sequence becomes longer and forgetting is typically more severe.
In contrast, our method maintains robust performance under fine-grained incremental scenarios, indicating that the proposed dynamic expert expansion and pruning mechanism is effective for handling long class-incremental sequences.

\begin{table}[htbp]
\centering
\caption{Comparison of state-of-the-art CL methods on the CIFAR100 and Tiny-ImageNet in terms of ``Avg.'' and ``Last'' scores (\%).}
\label{tab:cil_results}
\begin{small}
\setlength{\tabcolsep}{3pt}
\begin{tabular}{ll|cc|cc|cc |cc|cc|cc}
    \toprule
     & 
     & \multicolumn{6}{c|}{\textbf{CIFAR100}} 
     & \multicolumn{6}{c}{\textbf{Tiny-ImageNet}} \\
    \cmidrule(l){3-8} 
    \cmidrule(l){9-14}
    \textbf{Method} & 
    & \multicolumn{2}{c}{10 step} 
    & \multicolumn{2}{c}{20 step} 
    & \multicolumn{2}{c|}{50 step} 
    & \multicolumn{2}{c}{5 step} 
    & \multicolumn{2}{c}{10 step} 
    & \multicolumn{2}{c}{20 step} \\
    \cmidrule(l){3-14}
     & 
    & Avg. & Last 
    & Avg. & Last 
    & Avg. & Last 
    & Avg. & Last 
    & Avg. & Last 
    & Avg. & Last \\
    \midrule
    CLIP Zero-shot 
    & & 74.47 & 65.92 & 75.20 & 65.74 & 75.67 & 65.94
    & 69.62 & 65.30 & 69.55 & 65.59 & 69.49 & 65.30 \\
    Continual-FT 
    & & 65.46 & 53.23 & 59.69 & 43.13 & 39.23 & 18.89
    & 61.54 & 46.66 & 57.05 & 41.54 & 54.62 & 44.55 \\
    \midrule
    LwF~\cite{li2017learning}
    & & 65.86 & 48.04 & 60.64 & 40.56 & 47.69 & 32.90
    & 60.97 & 48.77 & 57.60 & 44.00 & 54.79 & 42.26 \\
    iCaRL~\cite{rebuffi2017icarl}
    & & 79.35 & 70.97 & 73.32 & 64.55 & 71.28 & 59.07
    & 72.05 & 70.39 & 73.43 & 67.95 & 69.65 & 64.68 \\
    LwF-VR~\cite{dhar2019learning}
    & & 78.81 & 70.75 & 74.54 & 63.54 & 71.02 & 59.45
    & 77.56 & 70.89 & 74.12 & 67.05 & 69.94 & 63.89 \\
    \midrule
    ZSCL~\cite{zheng2023preventing}  
    & & 82.15 & 73.65 & 80.39 & 69.58 & 79.92 & 67.36
    & 80.27 & 73.57 & 78.61 & 71.62 & 77.18 & 68.30 \\
    MoE-Adapters~\cite{yu2024boosting}  
    & & 85.21 & 77.52 & 83.72 & 76.20 & 83.60 & 75.24
    & 81.12 & 76.81 & 80.23 & 76.35 & 79.96 & 75.77 \\
    \rowcolor{lightblue!80}
    \textbf{Ours} 
    & & \textbf{86.14} & \textbf{78.59} 
    & \textbf{84.83} & \textbf{76.70} 
    & \textbf{84.21} & \textbf{76.22}
    & \textbf{82.36} & \textbf{78.42} 
    & \textbf{81.54} & \textbf{77.90} 
    & \textbf{80.93} & \textbf{77.11} \\
    \bottomrule
\end{tabular}
\end{small}
\end{table}

\paragraph{Expert Evolution Analysis.}
To further analyze how SCEE adapts the model capacity in the CIL setting, we report the average number of experts in each adapter layer on CIFAR100 with 10-task class-incremental learning.
As shown in \autoref{tab:cil_expert_number}, the number of experts is larger in shallow and middle layers, while deeper layers tend to use fewer experts.
This suggests that earlier layers require more diverse expert capacity to capture class-specific low- and mid-level visual variations introduced by new incremental tasks.
In contrast, deeper layers mainly encode higher-level semantic representations, where the pretrained CLIP features are more stable and transferable.
On average, only 2.48 experts are retained across layers, indicating that SCEE can adaptively allocate model capacity according to layer-wise demands while keeping the overall expert structure compact.

\begin{table}[ht]
\centering
\caption{
Layer-wise number of experts on CIFAR100 with 10-step class-incremental learning.
}
\label{tab:cil_expert_number}
\setlength{\tabcolsep}{6pt}
\renewcommand{\arraystretch}{1.15}
\small
\begin{tabular}{c|ccccccccccc|c}
\toprule
Layer 
& 0 & 1 & 2 & 3 & 4 & 5 & 6 & 7 & 8 & 9 & 10 & Avg. \\
\midrule
\#Experts 
& 3.92 & 3.18 & 3.67 & 2.41 & 2.86 & 2.24 & 2.08 & 1.95 & 1.63 & 1.84 & 1.47 & 2.48 \\
\bottomrule
\end{tabular}
\end{table}

\section{Results on DIL Benchmarks}
\label{sec:supp-dil-results}

\paragraph{Benchmark Description.}
We evaluate our method under the domain-incremental learning (DIL) setting on the DomainNet benchmark~\cite{peng2019moment}.
DomainNet contains images from multiple visually distinct domains, making it a challenging benchmark for evaluating continual learning under large domain shifts.
Following standard protocols in prior DIL and continual domain adaptation work~\cite{peng2019moment}, the model is trained sequentially over six domains, where each task corresponds to one visual domain.
The incremental learning follows the fixed domain order: ClipArt, Infograph, Painting, QuickDraw, Real, and Sketch.
At each incremental stage, the model is trained only on the current domain and evaluated on all previously seen domains.
We report the average accuracy over all incremental stages (``Avg.''), the final accuracy after learning all domains (``Last''), and the transfer performance to future domains (``Transfer''), following common practice in domain-incremental learning.

\paragraph{Experimental Setup.}
The experimental setup for DIL follows the same training protocol as the MTIL experiments.
All CLIP-based methods are implemented using the CLIP ViT-B/16 backbone~\cite{radford2021learning}, and the same optimization settings, adapter architecture, and evaluation protocol are adopted for fair comparison.
Different from standard CIL, where the task identity is unavailable and class boundaries change over time, DIL mainly focuses on sequential domain shifts under a shared label space.
Therefore, during inference, our method employs PGES to identify the most likely domain/task for each input sample and selects the corresponding routing strategy.
During training, SCEE is used to dynamically manage the expert structure by expanding experts for newly observed domain distributions and pruning redundant experts when necessary.
This enables the model to adaptively allocate capacity to domain-specific variations while preserving compact and transferable representations across domains.

\begin{figure}[t]
  \begin{center}
    \includegraphics[width=1.0\columnwidth]{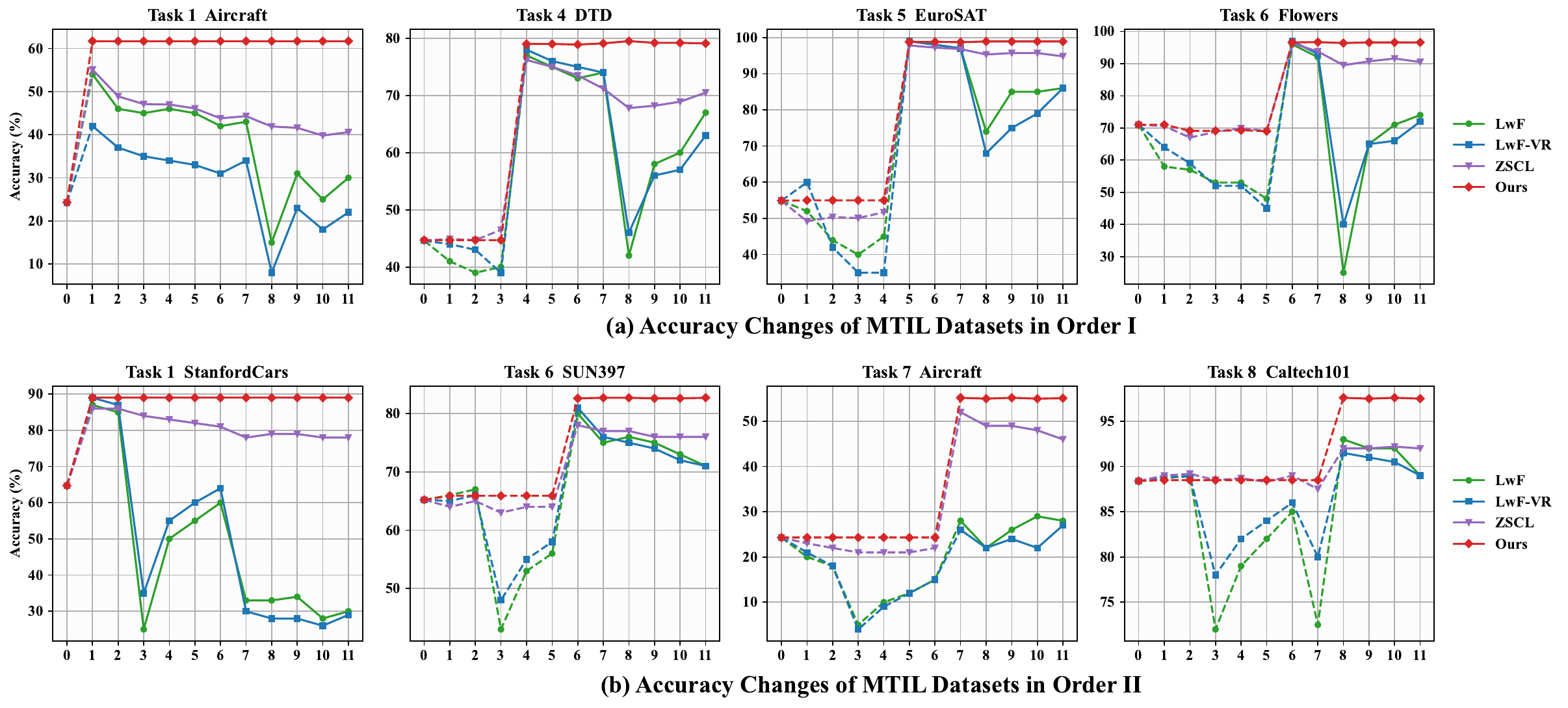}
    \caption{
    Illustration of the classification accuracy changes as tasks are being learned on the MTIL in two orders. The dashed lines represent the results of zero-shot predictions for an unlearned task. At task 0, the initial CLIP model’s zero-shot accuracy is evaluated.
    }
    \label{fig:mtl_accuracy_orderI_orderII}
  \end{center}
\end{figure}

\paragraph{Compared Methods.}
We compare our approach with representative state-of-the-art methods for domain-incremental learning.
These include knowledge distillation-based methods such as LwF~\cite{li2017learning} and LwF-VR~\cite{dhar2019learning}, optimization-based methods such as WiSE-FT~\cite{wortsman2022robust}, and recent CLIP-based continual learning approaches including ZSCL~\cite{zheng2023preventing} and MoE-Adapters~\cite{yu2024boosting}.
We also report the performance of CLIP zero-shot inference, denoted as Zero-shot~\cite{radford2021learning}, and continual fine-tuning without any forgetting mitigation strategy, denoted as Full Fine-tune, as reference baselines.
For fair comparison, all CLIP-based methods, including ours, follow the same backbone and training protocol unless otherwise specified.

\begin{wrapfigure}{r}{0.5\textwidth}
\vspace{-1.0em}
\centering
\captionof{table}{
Comparison of state-of-the-art methods on the DomainNet dataset in terms of ``Transfer'', ``Avg.'', and ``Last'' scores (\%).
}
\label{tab:DIL}
\vspace{-0.5em}

\setlength{\tabcolsep}{5.5pt}
\renewcommand{\arraystretch}{1.05}
\small
\begin{tabular}{lccc}
    \toprule
    \textbf{Method} & \textbf{Transfer} & \textbf{Avg.} & \textbf{Last} \\
    \midrule
    Zero-shot        & 57.86 & 60.79 & 60.79 \\
    Full Fine-tune   & 51.25 & 59.03 & 56.09 \\
    \midrule
    LwF~\cite{li2017learning}    & 52.54 & 63.14 & 66.13 \\
    LwF-VR~\cite{dhar2019learning} & 53.57 & 64.03 & 66.95 \\
    WiSE-FT~\cite{ding2022don} & 54.02 & 64.57 & 67.95 \\
    \midrule
    ZSCL~\cite{zheng2023preventing} & 55.74 & 65.13 & 68.73 \\
    MoE-Adapters~\cite{yu2024boosting}  & 56.03 & 66.22 & 72.50 \\
    \midrule
    \rowcolor{lightblue!80}
    \textbf{Ours} 
    & \textbf{57.00} 
    & \textbf{68.60} 
    & \textbf{75.20} \\
    \bottomrule
\end{tabular}
\vspace{-1.0em}
\end{wrapfigure}

\paragraph{Result Analysis.}
As shown in \autoref{tab:DIL}, our method consistently outperforms all compared approaches on the DomainNet benchmark across all evaluation metrics.
Compared with prior CLIP-based continual learning methods, our approach achieves notable improvements in both ``Avg.'' and ``Last'' scores, indicating stronger domain-level knowledge retention while maintaining effective adaptation to new domains.
In particular, the improvement in ``Last'' demonstrates that our method suffers less from accumulated forgetting after learning all domains.
Meanwhile, the competitive ``Transfer'' score suggests that our method largely preserves the zero-shot generalization capability of CLIP on unseen domains.
These results benefit from the combination of PGES and SCEE: PGES provides reliable task-aware routing during inference, while SCEE adaptively adjusts the expert capacity to accommodate new domain distributions.
Overall, these results confirm the effectiveness of our approach in the challenging domain-incremental learning setting, where substantial domain shifts intensify the stability--plasticity trade-off.


\begin{table*}[t]
    \centering
    \setlength{\tabcolsep}{1pt}
    \caption{Comparison with state-of-the-art methods on MTIL benchmark (\textbf{Order II}) in terms of ``Transfer'', ``Avg'', and ``Last'' scores (\%).We label the best methods with \textbf{bold} styles. The top block indicates the upper-bound solutions to adapt the CLIP on each task.}
    \label{tab:MTIL-order-II}
    \begin{small}
    \begin{tabular}{
        l
        *{11}{>{\centering\arraybackslash}p{0.8cm}}
        >{\centering\arraybackslash}p{1.4cm}
        }
        \toprule
        \textbf{Method} &
        \rotatebox{45}{Cars} &
        \rotatebox{45}{Food} &
        \rotatebox{45}{MNIST} &
        \rotatebox{45}{OxfordPet} &
        \rotatebox{45}{Flowers} &
        \rotatebox{45}{SUN397} &
        \rotatebox{45}{Aircraft} &
        \rotatebox{45}{Caltech101} &
        \rotatebox{45}{DTD} &
        \rotatebox{45}{EuroSAT} &
        \rotatebox{45}{CIFAR100} &
        Average \\
        \midrule

        \quad Zero-shot  & 64.7 & 88.5 & 59.4 & 89.0 & 71.0 & 65.2 & 24.3 & 88.4 & 44.6 & 54.9 & 68.2 & 65.4 \\
        \quad Fine-tune        & 89.6 & 92.7 & 99.6 & 94.7 & 97.5 & 81.8 & 62.0 & 95.1 & 79.5 & 98.9 & 89.6 & 89.2 \\
        \midrule

        \textbf{Transfer} & & & & & & & & & & & & \\
        \quad LwF~\cite{li2017learning}      & -- & 87.8 & 58.5 & 71.9 & 46.6 & 57.3 & 12.8 & 81.4 & 34.5 & 34.5 & 46.8 & 53.2 \\
        \quad iCaRL~\cite{rebuffi2017icarl}    & -- & 86.1 & 51.8 & 67.6 & 50.4 & 57.9 & 11.0 & 72.3 & 31.2 & 32.7 & 48.1 & 50.9 \\
        \quad WiSE-FT~\cite{wortsman2022robust}   & -- & 87.2 & 57.6 & 67.0 & 45.0 & 54.0 & 12.9 & 78.6 & 35.5 & 28.4 & 44.3 & 51.1 \\
        \quad ZSCL~\cite{zheng2023preventing}     & -- & 88.3 & 57.5 & 84.7 & 68.1 & 64.8 & 21.1 & 88.2 & 45.3 & 55.2 & \textbf{68.2} & 64.1 \\
        \quad MoE-Adapters~\cite{yu2024boosting}        & -- & \textbf{88.8} & 59.5 & 89.1 & 69.9 & 64.4 & 18.1 & 86.9 & 43.7 & 54.6 & \textbf{68.2} & 64.3 \\ 
        \quad LEBA~\cite{Gu2025gulearn} & -- & 88.7 & \textbf{60.2} & \textbf{89.3} & \textbf{71.1} & 65.1 & 18.4 & \textbf{88.5} & \textbf{45.9} & \textbf{55.3} & 68.1 & 65.1 \\
        \rowcolor{lightblue!80}
        \quad \textbf{Ours}     & -- & 88.5 & 58.9 & 89.1 & 71.0 & \textbf{65.9} & \textbf{24.4} & \textbf{88.5} & 44.7 & 55.0 & \textbf{68.2} & \textbf{65.4}\\
        \midrule

        \textbf{Avg.} & & & & & & & & & & & & \\
        \quad LwF~\cite{li2017learning}        & 49.0 & 77.0 & 92.1 & 85.9 & 66.5 & 67.2 & 20.9 & 84.7 & 44.6 & 45.5 & 50.5 & 62.2 \\
        \quad iCaRL~\cite{rebuffi2017icarl}      & 52.0 & 75.9 & 77.4 & 74.6 & 58.4 & 59.3 & 11.7 & 79.6 & 42.1 & 43.2 & 51.7 & 56.9 \\
        \quad WiSE-FT~\cite{ding2022don}  & 52.6 & 79.3 & 91.9 & 83.9 & 63.4 & 65.2 & 23.3 & 83.7 & 45.4 & 40.0 & 48.2 & 61.5 \\
        \quad ZSCL~\cite{zheng2023preventing}     & 81.7 & 91.3 & 91.1 & 91.0 & 82.9 & 72.5 & 33.6 & 89.7 & 53.3 & 62.8 & 69.9 & 74.5 \\
        \quad MoE-Adapters~\cite{yu2024boosting}       & 84.9 & 89.9 & 89.3 & 91.4 & 86.2 & 72.2 & 33.4 & 89.4 & 53.3 & 61.4 & 69.9 & 74.7 \\

        \quad LEBA~\cite{Gu2025gulearn} & 86.0 & 88.9 & 92.1 & \textbf{91.9} & 87.2 & 72.8 & 33.9 & 90.9 & \textbf{54.7} & 62.7 & \textbf{70.1} & 75.6 \\
        \rowcolor{lightblue!80}
        \quad \textbf{Ours}     & \textbf{89.0} & \textbf{92.3} & \textbf{92.2} & 91.7 & \textbf{87.4} & \textbf{75.0} & \textbf{38.3} & \textbf{91.8} & 54.0 & \textbf{63.0} & \textbf{70.1} & \textbf{76.8}\\
        \midrule

        \textbf{Last} & & & & & & & & & & & & \\
        \quad LwF~\cite{li2017learning}         & 34.6 & 69.6 & 99.3 & 88.7 & 61.1 & 72.5 & 32.5 & 88.1 & 65.6 & 90.9 & 87.9 & 71.9 \\
        \quad iCaRL~\cite{rebuffi2017icarl}      & 46.0 & 81.5 & 91.3 & 82.8 & 66.5 & 72.2 & 16.3 & 91.6 & 68.1 & 83.2 & 87.8 & 71.6 \\
        \quad WiSE-FT~\cite{ding2022don}    & 35.6 & 76.9 & 99.5 & 89.1 & 62.1 & 71.8 & 27.8 & 90.8 & 67.0 & 85.6 & 87.6 & 72.2 \\
        \quad ZSCL~\cite{zheng2023preventing}       & 78.2 & 91.1 & 97.6 & 92.5 & 87.4 & 78.2 & 45.0 & 92.3 & 72.7 & 96.2 & 86.3 & 83.4 \\
        \quad MoE-Adapters~\cite{yu2024boosting}        & 84.1 & 88.5 & 94.0 & 91.8 & 94.1 & 77.8 & 50.4 & 93.3 & 77.1 & 87.7 & 86.6 & 84.1 \\
        \quad LEBA~\cite{Gu2025gulearn} & 86.2 & 88.9 & 99.2 & \textbf{93.0} & 96.5 & 79.2 & 50.1 & 95.2 & 78.2 & 95.9 & 88.1 & 86.4 \\
        \rowcolor{lightblue!80}
        \quad \textbf{Ours}  & \textbf{89.0} & \textbf{92.6} & \textbf{99.6} & 92.7 & \textbf{96.7} & \textbf{82.7} & \textbf{55.1} & \textbf{97.5} & \textbf{79.0} & \textbf{98.8} & \textbf{88.7} & \textbf{88.4}\\
\bottomrule
\end{tabular}
\end{small}
\end{table*}

\begin{table*}[t]
    \centering
    \setlength{\tabcolsep}{1pt}
    \caption{Comparison with state-of-the-art methods on few-shot MTIL benchmark (\textbf{Order~II}) in terms of ``Transfer'', ``Avg.'', and ``Last'' scores (\%). We label the best methods with \textbf{bold} styles. The top block indicates the upper-bound solutions to adapt the CLIP on each task.}
    \label{tab:FS-MTIL-order-II}
    \begin{small}
        \begin{tabular}{
        l
        *{11}{>{\centering\arraybackslash}p{0.8cm}}
        >{\centering\arraybackslash}p{1.4cm}
        }
        \toprule
        \textbf{Method} &
        \rotatebox{45}{Cars} &
        \rotatebox{45}{Food} &
        \rotatebox{45}{MNIST} &
        \rotatebox{45}{OxfordPet} &
        \rotatebox{45}{Flowers} &
        \rotatebox{45}{SUN397} &
        \rotatebox{45}{Aircraft} &
        \rotatebox{45}{Caltech101} &
        \rotatebox{45}{DTD} &
        \rotatebox{45}{EuroSAT} &
        \rotatebox{45}{CIFAR100} &
        Average \\
        \midrule
        \quad Zero-shot
        & 64.7 & 88.5 & 59.4 & 89.0 & 71.0 & 65.2 & 24.3 & 88.4 & 44.6 & 54.9 & 68.2 & 65.3 \\
        \quad 5-shot Full Fine-tune
        & 65.4 & 83.3 & 96.6 & 84.9 & 92.9 & 71.3 & 30.6 & 93.5 & 65.1 & 91.7 & 76.8 & 77.5 \\
        \midrule
        \midrule

        \textbf{Transfer} & & & & & & & & & & & & \\
        \quad LwF~\cite{li2017learning}
        & -- & 64.2 & 59.1 & 68.1 & 38.4 & 54.9 & 6.7 & 78.0 & 35.5 & 33.5 & 47.4 & 48.6 \\
        \quad LwF-VR~\cite{ding2022don}
        & -- & 80.1 & 55.4 & 77.7 & 50.4 & 61.4 & 9.1 & 83.5 & 40.1 & 31.5 & 54.8 & 54.4 \\
        \quad WiSE-FT~\cite{wortsman2022robust}
        & -- & 77.3 & 60.0 & 76.9 & 54.2 & 58.0 & 11.1 & 81.8 & 37.6 & 31.7 & 48.1 & 53.7 \\
        \quad ZSCL~\cite{zheng2023preventing}
        & -- & 87.3 & \textbf{64.8} & 85.3 & 67.9 & 64.5 & 18.9 & 86.6 & 43.6 & 43.2 & 65.7 & 62.8  \\
        \quad MoE-Adapters~\cite{yu2024boosting}
        & -- & \textbf{88.8} & 59.5 & \textbf{89.1} & \textbf{71.2} & 65.3 & 18.2 & 87.9 & 44.2 & 54.6 & \textbf{68.2} & 64.7 \\
        \rowcolor{lightblue!80}
        \quad \textbf{Ours} 
        & -- & 88.5 & 58.9 & \textbf{89.1} & 71.0 & \textbf{65.9} & \textbf{24.4} & \textbf{88.5} & \textbf{44.7} & \textbf{55.0} & \textbf{68.2} & \textbf{65.4}\\
        \midrule

        \textbf{Avg.} & & & & & & & & & & & & \\
        \quad LwF~\cite{li2017learning}
        & 64.1 & 55.0 & 79.5 & 69.2 & 55.7 & 58.3 & 10.8 & 81.7 & 41.3 & 39.2 & 47.4 & 54.7 \\
        \quad LwF-VR~\cite{ding2022don}
        & 63.3 & 76.9 & 71.4 & 79.1 & 68.9 & 65.0 & 13.4 & 86.0 & 45.7 & 36.3 & 55.3 & 60.1 \\
        \quad WiSE-FT~\cite{wortsman2022robust}
        & 59.3 & 64.7 & 77.4 & 70.3 & 51.3 & 58.6 & 10.8 & 84.2 & 42.0 & 38.6 & 49.1 & 55.1 \\
        \quad ZSCL~\cite{zheng2023preventing}
        & 70.0 & 85.0 & 79.8 & 86.1 & 79.4 & 68.3 & 21.8 & 88.8 & 48.8 & 49.3 & 66.5 & 67.6 \\
        \quad MoE-Adapters~\cite{yu2024boosting}
        & 61.2 & 87.0 & \textbf{87.3} & 89.1 & 79.3 & 68.5 & 23.4 & \textbf{89.4} & \textbf{49.9} & \textbf{60.8} & 68.8 & 69.5 \\
        \rowcolor{lightblue!80}
        \quad \textbf{Ours}
        & \textbf{70.6} & \textbf{87.8} & 85.2 & \textbf{90.4} & \textbf{83.2} & \textbf{69.7} & \textbf{29.6} & \textbf{89.4} & 49.7 & \textbf{60.8} & \textbf{69.0} & \textbf{71.4}\\
        \midrule

        \textbf{Last} & & & & & & & & & & & & \\
        \quad LwF~\cite{li2017learning}
        & 57.1 & 40.1 & 84.1 & 58.1 & 50.5 & 57.6 & 14.3 & 87.9 & 54.7 & 64.0 & 47.0 & 56.8 \\
        \quad LwF-VR~\cite{ding2022don}
        & 57.3 & 70.1 & 72.1 & 74.6 & 71.9 & 65.8 & 17.4 & 89.5 & 60.0 & 56.0 & 60.2 & 63.5 \\
        \quad WiSE-FT~\cite{wortsman2022robust}
        & 48.1 & 47.7 & 66.9 & 59.8 & 25.0 & 56.1 & 7.4 & 88.5 & 52.2 & 66.8 & 59.4 & 51.8 \\
        \quad ZSCL~\cite{zheng2023preventing}
        & 67.4 & 82.7 & 78.7 & 85.7 & 81.3 & 71.2 & 25.0 & \textbf{92.5} & 62.0 & 72.2 & 74.4 & 71.8 \\
        \quad MoE-Adapters~\cite{yu2024boosting}
        & 59.4 & 87.0 & \textbf{91.8} & 89.0 & 84.1 & 71.9 & 29.4 & 91.4 & \textbf{64.2} & \textbf{88.8} & 75.0 & 75.7 \\
        \rowcolor{lightblue!80}
        \quad \textbf{Ours}
        & \textbf{70.5} & \textbf{87.7} & 91.1 & \textbf{90.9} & \textbf{90.4} & \textbf{72.9} & \textbf{35.8} & 91.0 & 62.6 & 86.8 & \textbf{77.0} & \textbf{77.9} \\
        \bottomrule
\end{tabular}
\end{small}
\end{table*}

\section{Detailed Results on MTIL Benchmark}
\label{sec:supp-MTIL-results}
\paragraph{Additional Benchmark Description.}
The MTIL benchmark consists of 11 datasets spanning diverse visual domains, designed to evaluate continual learning under the Multi-Task Incremental-Learning setting~\cite{zheng2023preventing}.
Following prior work~\cite{zheng2023preventing}, we evaluate two different task sequences to induce varying domain shifts.
\textbf{Order I} follows an alphabetical order of datasets: Aircraft~\cite{maji2013fine}, Caltech101~\cite{fei2004learning}, CIFAR100~\cite{krizhevsky2009learning}, DTD~\cite{cimpoi2014describing}, EuroSAT~\cite{helber2019eurosat}, Flowers~\cite{nilsback2008automated}, Food~\cite{bossard2014food}, MNIST~\cite{deng2012mnist}, OxfordPet~\cite{parkhi2012cats}, StanfordCars~\cite{krause20133d}, and SUN397~\cite{xiao2010sun}.
\textbf{Order II} adopts a randomly permuted order: StanfordCars~\cite{krause20133d}, Food~\cite{bossard2014food}, MNIST~\cite{deng2012mnist}, OxfordPet~\cite{parkhi2012cats}, Flowers~\cite{nilsback2008automated}, SUN397~\cite{xiao2010sun}, Aircraft~\cite{maji2013fine}, Caltech101~\cite{fei2004learning}, DTD~\cite{cimpoi2014describing}, EuroSAT~\cite{helber2019eurosat}, and CIFAR100~\cite{krizhevsky2009learning}.

\paragraph{Additional Metrics Formulation.}
We evaluate continual learning performance using Transfer, Avg., and Last.
Let $a_k^{(j)}$ denote the test accuracy on task $k$ after the model has been trained on the first $j$ tasks,
and let $K$ be the total number of tasks.
\begin{equation}
\mathrm{Transfer}_k
= \frac{1}{k-1} \sum_{j=1}^{k-1} a_k^{(j)},
\quad k = 2, \ldots, K .
\end{equation}
\begin{equation}
\mathrm{Avg}_k
= \frac{1}{K} \sum_{j=1}^{K} a_k^{(j)},
\quad k = 1, \ldots, K .
\end{equation}
\begin{equation}
\mathrm{Last}_k
= a_k^{(K)},
\quad k = 1, \ldots, K .
\end{equation}
The Transfer metric evaluates the preservation of zero-shot capability during continual learning by averaging the model’s performance on tasks prior to their explicit training.
The Last metric reports the final accuracy on each task after all training stages, reflecting the degree of knowledge retention.
The Avg. metric computes the mean accuracy across all training stages, summarizing the model’s overall performance throughout the continual learning process.

\paragraph{Detailed Results.}
\autoref{tab:MTIL-order-II} and \autoref{tab:FS-MTIL-order-II} report the detailed Transfer, Avg.\, and Last results for each dataset on the full-shot MTIL and few-shot MTIL benchmarks under Order~II.
Zero-shot denotes the zero-shot prediction performance of the initial CLIP model, while Fine-tune refers to training the model on each dataset independently without task interference.
Across both full-shot and few-shot MTIL Order~II benchmarks, our method achieves the best or highly competitive performance on most datasets under all three metrics, demonstrating its strong ability to preserve pretrained knowledge while effectively learning new tasks under challenging domain shifts.

To provide a more intuitive understanding of model behavior, we visualize the accuracy evolution of several representative tasks under both task orderings in \autoref{fig:mtl_accuracy_orderI_orderII}(a) and \autoref{fig:mtl_accuracy_orderI_orderII}(b), where dashed curves denote zero-shot performance before a task is encountered and solid curves indicate accuracy after task-specific training.
An ideal continual learning approach for vision--language models should exhibit an accuracy trajectory resembling a mirrored ``Z'' shape, maintaining high zero-shot accuracy prior to learning a task and experiencing minimal degradation after learning subsequent tasks, a pattern that our method consistently follows under both task sequences.
In contrast, most baseline methods suffer from pronounced accuracy fluctuations caused by catastrophic forgetting of both pretrained knowledge and previously learned tasks, a problem that becomes particularly severe under large domain shifts such as MNIST, corresponding to Task~8 in Order~I and Task~3 in Order~II, where naive fine-tuning and regularization-based approaches experience substantial performance collapse.
Overall, the strong performance of our method demonstrates its ability to effectively resist domain shifts and maintain stable learning behavior over longer task sequences.

\begin{figure}[htbp]
  \begin{center}
    \includegraphics[width=1.0\columnwidth]{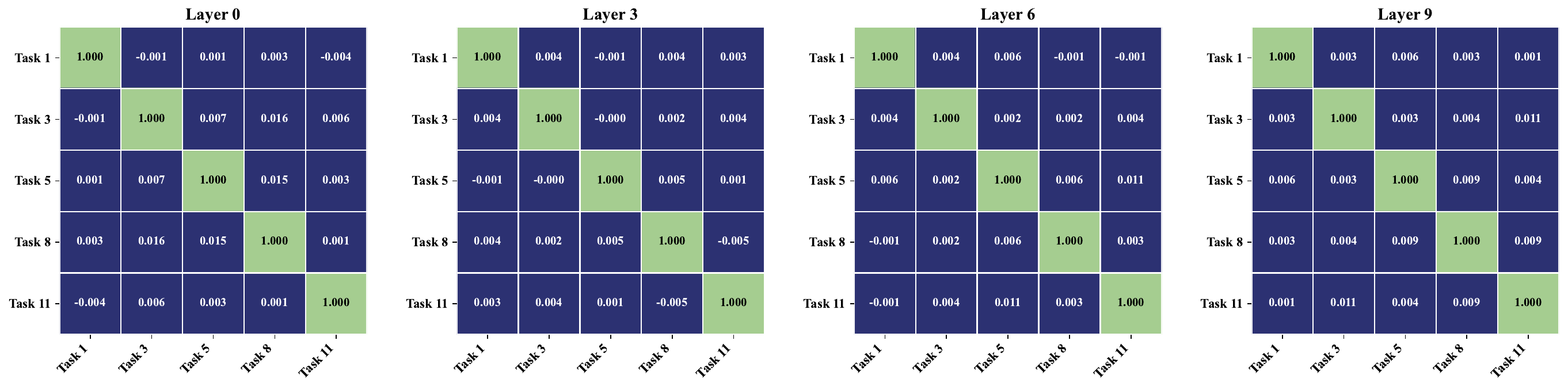}
    \caption{
    Layer-wise task-specific expert similarity matrix.
We record the cosine similarity between task-specific experts per layer on the MTIL Order I benchmark after training. The results show that the cosine similarity between experts is close to zero.
    }
    \label{fig:task_expert_similarity}
  \end{center}
\end{figure}

\section{Additional Experimental Results}
\label{sec:additional-experimetal-results}

\paragraph{Layer-wise Task-Specific Expert Similarity.}
To further investigate whether the proposed dynamic expert mechanism leads to redundant expert representations, we analyze the pairwise cosine similarity among task-specific experts across different MoE layers in the visual encoder. Specifically, for each incremental task, we randomly select one expert from the set of newly introduced experts associated with that task as its representative, and compute the cosine similarity between expert parameter vectors after training. \autoref{fig:task_expert_similarity} presents the layer-wise expert similarity matrices for selected MoE layers, with each matrix reporting the cosine similarity between experts associated with different tasks. By definition, the diagonal entries equal 1, while the off-diagonal entries remain consistently close to zero across all layers. These results indicate that the task-specific experts learned by our method are nearly orthogonal rather than redundant, suggesting that different experts capture distinct task-dependent representations. 

\begin{wrapfigure}{r}{0.55\textwidth}
\vspace{-1.0em}
\centering
\includegraphics[width=\linewidth]{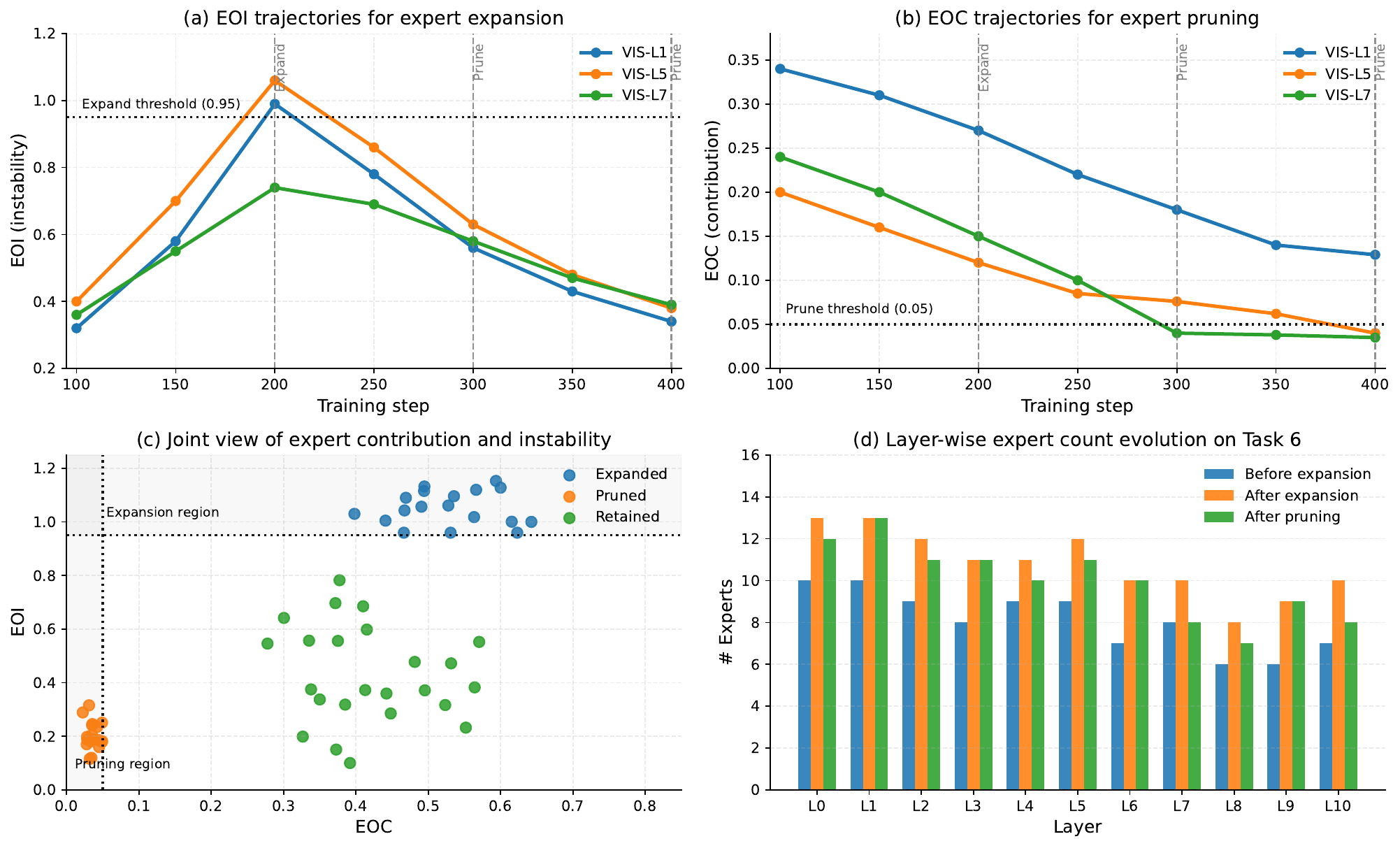}
\caption{
Analysis of expert evolution in SCEE on Task 6 (Flowers).
}
\label{fig:scee_evolution}
\vspace{-1.0em}
\end{wrapfigure}

\paragraph{Analysis of Expert Evolution in SCEE.}
To understand how SCEE dynamically controls the expert structure, we visualize the expert evolution process on Task 6 (Flowers), as shown in \autoref{fig:scee_evolution}. Subplot (a) shows the trajectory of Expert Optimization Instability (EOI), where layers L1 and L5 exceed the expansion threshold at step 200 and trigger expert expansion, while L7 remains below the threshold and does not introduce new experts. Subplot (b) presents the trajectory of Expert Optimization Contribution (EOC), showing that low-contribution experts are gradually pruned; specifically, L5 undergoes one pruning event, while L7 undergoes two pruning events. Subplot (c) further visualizes the joint distribution of EOC and EOI, where experts are clearly separated into expansion, retention, and pruning regions, verifying that expert evolution follows the proposed gradient-based criteria. Subplot (d) shows the layer-wise number of experts before expansion, after expansion, and after pruning. Shallow layers tend to expand more experts, whereas deeper layers are more likely to prune redundant ones, leading to a hierarchical expert structure. Notably, L7 keeps the same number of experts after expansion and pruning, indicating that it mainly relies on reusing existing experts. Overall, these results show that SCEE dynamically adjusts model capacity according to gradient statistics rather than static heuristics, enabling effective layer-wise expert allocation.

\paragraph{Analysis of PGES Routing Accuracy}
To further evaluate the reliability of the proposed PGES module in the MTIL setting, we report the task routing accuracy across the incremental learning process in \autoref{tab:pges_routing_accuracy}. 
Each row denotes the current training stage, and each column denotes the target task to be identified. 
For seen tasks, corresponding to the entries on and below the diagonal, PGES achieves nearly perfect routing accuracy in most cases.
The only noticeable decrease appears on DTD, where the routing accuracy is 87.5\% in several stages.
For unseen tasks, corresponding to the entries above the diagonal, most samples are correctly routed to the frozen CLIP branch rather than being assigned to existing task-specific experts.
Only minor leakage is observed, such as about 8\% of Flowers samples being routed to learned experts before Flowers is introduced.

\begin{table}[t]
\centering
\caption{
Routing accuracy of PGES in the MTIL setting.
Rows indicate the current training stage, while columns indicate target tasks.
Entries on and below the diagonal correspond to seen-task routing accuracy, and entries above the diagonal indicate the proportion of unseen-task samples routed to learned experts.
}
\label{tab:pges_routing_accuracy}
\setlength{\tabcolsep}{4.5pt}
\renewcommand{\arraystretch}{1.25}
\scriptsize
\begin{tabular}{lccccccccccc}
\toprule
Task 
& Aircraft & Caltech101 & CIFAR100 & DTD & EuroSAT & Flowers 
& Food & MNIST & OxfordPet & Cars & SUN397 \\
\midrule
Aircraft   
& 100.0 & 0 & 0 & 0 & 0 & 0 & 0 & 0 & 0 & 0 & 0 \\
Caltech101 
& 100.0 & 100.0 & 0 & 0 & 0 & 0 & 0 & 0 & 0 & 0 & 0 \\
CIFAR100   
& 100.0 & 100.0 & 100.0 & 0 & 0 & 0.08 & 0 & 0 & 0 & 0 & 0 \\
DTD        
& 100.0 & 100.0 & 100.0 & 87.5 & 0 & 0.08 & 0 & 0 & 0 & 0 & 0 \\
EuroSAT    
& 100.0 & 100.0 & 100.0 & 87.5 & 100.0 & 0.08 & 0 & 0 & 0 & 0 & 0 \\
Flowers    
& 100.0 & 100.0 & 100.0 & 87.5 & 100.0 & 100.0 & 0 & 0 & 0 & 0 & 0 \\
Food       
& 100.0 & 100.0 & 100.0 & 100.0 & 100.0 & 100.0 & 100.0 & 0 & 0 & 0 & 0 \\
MNIST      
& 100.0 & 100.0 & 100.0 & 100.0 & 100.0 & 100.0 & 100.0 & 100.0 & 0 & 0 & 0 \\
OxfordPet  
& 100.0 & 100.0 & 100.0 & 100.0 & 100.0 & 100.0 & 100.0 & 100.0 & 100.0 & 0 & 0 \\
Cars       
& 100.0 & 100.0 & 100.0 & 100.0 & 100.0 & 100.0 & 100.0 & 100.0 & 100.0 & 100.0 & 0 \\
SUN397     
& 100.0 & 100.0 & 100.0 & 100.0 & 100.0 & 100.0 & 100.0 & 100.0 & 100.0 & 100.0 & 100.0 \\
\bottomrule
\end{tabular}
\end{table}

The impact of the small amount of misrouting is limited.
For samples from seen tasks that are misrouted to the frozen CLIP branch, such as DTD, the model can still benefit from CLIP's strong zero-shot capability, leading to only about a 1\% performance drop.
For unseen tasks that are incorrectly assigned to existing experts, such as Flowers before its introduction, the leakage ratio is very small and introduces only about a 2\% degradation.
Overall, PGES provides highly reliable task identification in MTIL, and its rare routing errors have negligible influence on the final performance.

\section{Algorithm Procedure}
Algorithms~\ref{alg:training_dimoe} and~\ref{alg:inference_dimoe} detail the training and inference procedures of the proposed DIMoE-Adapters framework. 
During training, we sequentially learn task-specific routers and experts while maintaining per-task distribution prototypes $(\mu_{(t,k)}, \Sigma_{(t,k)})$ computed from frozen CLIP image features. 
At inference time, the learned task prototypes are used for task identification by evaluating the confidence scores of a test sample. 
If the maximum confidence exceeds a predefined threshold, the corresponding task-specific router is activated and the associated experts are injected into the frozen CLIP backbone for prediction; 
otherwise, the model falls back to the frozen CLIP to preserve zero-shot generalization.

\begin{algorithm}[t]
\caption{Training Procedure of DIMoE-Adapters}
\label{alg:training_dimoe}
\begin{algorithmic}[1]

\REQUIRE Sequential tasks $\{\tau^{(t)}\}_{t=1}^{T}$ with datasets 
$\mathcal{D}^{(t)}$; frozen CLIP image encoder $f$;\\
Top-$p$ threshold $p_0$; expansion threshold $\gamma_{\text{expand}}$; 
pruning thresholds $\gamma_{\text{prune}}$; \\
learning rate $\eta$; evolution window $W$; interval $\Delta$; max iterations $I_{\max}$.

\FOR{$t = 1$ to $T$}

    \FOR{$j = 1$ to $N_t$}
        \STATE Extract image feature $f(x_j^{(t)})$
        \STATE $\mathcal{B}_{\text{feat}} \leftarrow \mathcal{B}_{\text{feat}} \cup \{f(x_j^{(t)})\}$
    \ENDFOR

    \STATE Compute task prototype $(\mu_{(t,k)}, \Sigma_{(t,k)})$ from $\mathcal{B}_{\text{feat}}$
    \hfill $\triangleright$ Eq.~(\ref{eq:task_Prototype})
    
    \STATE Initialize router $R^{(t)}$
    \STATE Initialize new experts $\mathcal{E}_{\text{new}}^{(t)}$
    \IF{$t > 1$}
        \STATE Freeze $\{R^{(1)}, \dots, R^{(t-1)}\}$ and $\mathcal{E}^{(t-1)}$
        \STATE $\mathcal{E}^{(t)} \leftarrow \mathcal{E}^{(t-1)} \cup \mathcal{E}_{\text{new}}^{(t)}$
    \ELSE
        \STATE $\mathcal{E}^{(t)} \leftarrow \mathcal{E}_{\text{new}}^{(t)}$
    \ENDIF

    \FOR{$\ell = 1$ to $I_{\max}$}
        \STATE Sample mini-batch $\mathcal{B} \sim \mathcal{D}^{(t)}$
        \STATE Forward image and text encoders with DIMoE-Adapters
        \STATE Update $R^{(t)}$ and $\mathcal{E}_{\text{new}}^{(t)}$ by minimizing $\mathcal{L}$
    
        \STATE Accumulate gradients for experts
        \STATE Accumulate expert usage counts
        
        \IF{$\ell \bmod \Delta = 0$}
            \STATE Compute $\{I_i, V_i\}$ from accumulated gradients
            \hfill $\triangleright$ Eq.~(\ref{eq:I_curr}) ~(\ref{eq:V_curr})
            
            \STATE Compute activation frequency $f_i$ by normalizing usage counts
            
            \IF{$\ell \in W$}
                \STATE Perform expert pruning and expansion for $\mathcal{E}_{\text{new}}^{(t)}$
                \hfill $\triangleright$ Eq.~(\ref{eq:prune}) ~(\ref{eq:expand})
            \ENDIF
            
            \STATE Update historical adaptive baselines 
            \hfill $\triangleright$ Eq.~(\ref{eq:H_I_V}) 
            
            \STATE Reset accumulated gradients and usage counts
        \ENDIF
    \ENDFOR
\ENDFOR

\end{algorithmic}
\end{algorithm}

\begin{algorithm}[ht]
\caption{Inference Procedure of DIMoE-Adapters}
\label{alg:inference_dimoe}
\begin{algorithmic}[1]

\REQUIRE test sample $x$; frozen CLIP image encoder $f$; \\
task prototypes $\{(\mu_{(i,k)}, \Sigma_{(i,k)})\}_{i=1}^{t}$; 
task-specific routers $\{R^{(i)}\}_{i=1}^{t}$; expert pools $\mathcal{E}^{(t)}$; 
confidence threshold $\delta$.

\STATE Compute image feature $z = f(x)$

\STATE Compute task scores $\{S_i(z)\}_{i=1}^{t}$
\hfill $\triangleright$ Eq.~(\ref{eq:Inference}) 

\STATE $\hat{t} \leftarrow \arg\max_i S_i(z)$, \quad $S_{\max} \leftarrow S_{\hat{t}}$

\IF{$S_{\max} \ge \delta$}
    \STATE Activate router $R^{(\hat{t})}$
    \STATE Forward image and text encoders with DIMoE-Adapters
\ELSE
    \STATE Forward image and text encoders with frozen CLIP
\ENDIF
\end{algorithmic}
\end{algorithm}


\end{document}